\documentclass[11pt]{article}

\usepackage[final]{acl}

\usepackage{times}
\usepackage{latexsym}
\usepackage{subcaption}
\usepackage{algorithm}
\usepackage{algpseudocode}
\usepackage[table]{xcolor}
\usepackage{multirow}
\usepackage{booktabs}
\usepackage{makecell}
\usepackage{amsmath}
\usepackage{amssymb}
\usepackage{mathtools}
\usepackage{amsthm}

\newcommand{\yj}[1]{{\color{black}#1}}

\usepackage[T1]{fontenc}

\usepackage[utf8]{inputenc}

\usepackage{microtype}

\usepackage{inconsolata}

\usepackage{graphicx}

\title{SAGE: Sink-Aware Guided Emphasis for Visual Grounding in Vision-Language Decoders}

\author{
  \textbf{Jeonghyo Song}\textsuperscript{*},
  \textbf{YoungJoon Yoo}\textsuperscript{\dag}
\\
  Department of Artificial Intelligence, Chung-Ang University
\\
  \texttt{\{thd9592s, yjyoo3312\}@cau.ac.kr}
\\
  \small{\textsuperscript{*}First author. \textsuperscript{\dag}Corresponding author.}
}

\begin{document}
\maketitle

\begin{abstract}
Recent large vision-language models (VLMs) pair a visual encoder with a large language model (LLM) and perform well on diverse image--text tasks, yet their reliability is often limited by decoder attention pathologies that suppress visual evidence and exacerbate hallucinations.
In this paper, we revisit visual attention sinks and uncover a structured, layer-dependent behavior: across prompts, early and late decoder layers exhibit prompt-invariant attention collapse onto the same few image regions, which we term PIS (Prompt-Invariant Sinks), whereas mid layers become prompt-conditioned and drive vision-language alignment.
This split suggests that treating sinks as a uniform effect is incomplete. Building on this insight, we propose SAGE (Sink-Aware Guided Emphasis), a lightweight intervention that steers decoder attention away from PIS and toward query-dependent regions of interest (ROIs) using token-aligned ROI masks derived from standard vision backbones such as CLIP, ViT, and DINOv3.
Evaluated on diverse vision-encoder + decoder-only LLM VLM families, SAGE improves visual grounding, reduces hallucinations, and yields consistent gains across public downstream vision-language benchmarks, including fine-grained visual discrimination settings where localized evidence is crucial, when instantiated with backbone-derived ROI masks.
\end{abstract}

\section{Introduction}
\label{introduction}

Large vision–language models (VLMs)~\cite{liu2023visual,liu2024improved,li2023blip,wang2024qwen2,yang2025qwen3} have become a strong foundation for integrating visual perception with language reasoning. 
By connecting a visual encoder to a large language model (LLM)~\cite{achiam2023gpt,touvron2023llama,minaee2024large} decoder via lightweight adapters~\cite{wan2023efficient,van2025survey}, recent VLMs support a broad spectrum of applications, including image captioning, detailed image understanding, OCR, object localization, and multi-turn visual question answering within a unified interface. 
Despite these capabilities, reliability remains a central concern. 
VLMs may overlook subtle but decisive visual evidence, struggle with fine-grained recognition~\cite{tong2024eyes,kim2024finer,liao2025can}, and produce plausible yet incorrect statements that contradict the image, a behavior commonly referred to as hallucination~\cite{li2023evaluating,huang2024visual,chen2025survey,liu2024survey}.

\yj{Prior analyses suggest that these failures are not solely due to the visual encoder, but also to the decoder’s attention dynamics over visual tokens~\cite{kang2025see}. 
In particular, self-attention can concentrate on a small subset of tokens that consistently attracts an outsized share of the attention mass. 
In multimodal decoders, such visual attention sinks can dominate the attention budget and pull attention away from query-relevant image regions.
This reduces precise visual grounding, especially for questions that rely on subtle, localized evidence.}

\begin{figure*}[t]
\centering
\includegraphics[width=1\textwidth]{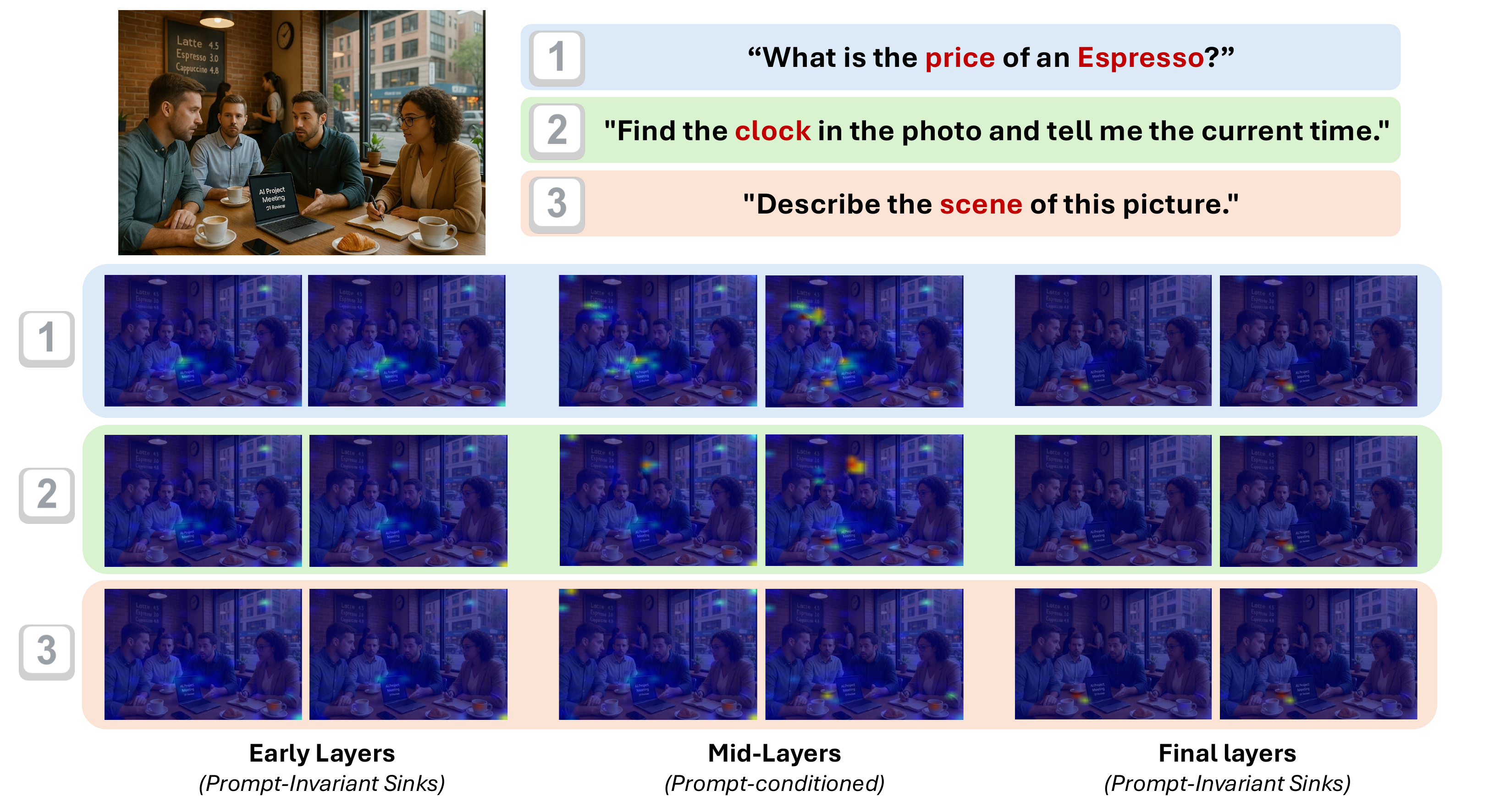} 
\vspace{-7mm}
\caption{\textbf{Prompt-Invariant Sinks and prompt-conditioned mid-layer attention in a VLM decoder.}
We visualize decoder attention over visual tokens for three different prompts (rows) at representative early, mid, and final decoder layers (columns).
Early and final layers exhibit prompt-invariant attention collapse onto nearly identical image regions across prompts, forming Prompt-Invariant Sinks.
In contrast, mid-layer attention is prompt-conditioned: it shifts to the menu board when asking for the espresso price (Prompt 1), to the wall clock when asking for the current time (Prompt 2), and becomes broader for holistic scene description (Prompt 3).
Warmer colors indicate higher attention.}
\label{fig:decoder_fig}
\end{figure*}

\yj{In this paper, we re-examine decoder-side visual attention sinks and identify a structured, layer-dependent behavior. 
This behavior is easy to miss when attention is aggregated across layers rather than analyzed layer by layer.}
Figure~\ref{fig:decoder_fig} provides a motivating example. Across three substantially different prompts, early decoder layers and final decoder layers repeatedly assign high attention to nearly identical image locations, even though the question changes. 
We call these stable, prompt-agnostic regions Prompt-Invariant Sinks. 
In contrast, the mid layers exhibit the only consistently prompt-responsive behavior.
When the query asks for the `espresso price', mid-layer attention shifts toward the menu board. When the query asks for the current time, mid-layer attention moves toward the clock. When the prompt requests a scene description, mid-layer attention becomes broader and better aligned with semantically relevant regions. 
These observations reveal a dual role of the decoder stack: attention is largely static near the beginning and the end, while meaningful vision–language alignment emerges primarily in the middle layers.

This layer-wise structure has two implications. First, treating visual attention sinks as a single uniform pathology is incomplete, because the decoder simultaneously contains components that are insensitive to the query and components that must perform alignment. Second, Prompt-Invariant Sinks offer a concrete target for intervention. If attention can be discouraged from collapsing onto these static sinks and instead encouraged to concentrate on query-dependent regions of interest, VLMs may become more faithful to the image and more effective at fine-grained reasoning.

Motivated by these observations, we introduce SAGE (Sink-Aware Guided Emphasis) as a lightweight and controlled method to establish where and how decoder-side attention interventions matter in encoder--decoder VLMs such as widely-used Qwen-VL~\cite{wang2024qwen2,yang2025qwen3}, LLaVA~\cite{liu2023visual,liu2024improved,li2025llava}-style, InternVL~\cite{zhu2025internvl3,chen2024internvl} and NVILA~\cite{liu2025nvila} architectures.
The core contribution of this work is the layer-dependent structure of visual attention sinks and alignment, and SAGE serves to make this effect measurable and reproducible.
Concretely, SAGE steers decoder attention away from Prompt-Invariant Sinks and toward query-dependent regions of interest (ROIs) during generation.
In our implementation, ROI masks are derived from standard pretrained vision backbones such as CLIP, ViT, and DINOv3, providing a consistent way to construct token-aligned prompt-conditioned interventions across models.
With SAGE, we sweep over the intervention layers and show that simple ROI-guided edits at specific decoder layers yield reliable gains in grounding and downstream performance.

Our contributions are summarized as threefold:
\begin{itemize}
    \item We identify Prompt-Invariant Sinks, showing that early and final decoder layers exhibit prompt-agnostic attention concentration on fixed image regions across diverse queries, as illustrated in Figure~\ref{fig:decoder_fig}.
    \item We reveal a layer-wise duality in VLM decoders, where mid layers are primarily responsible for prompt-responsive vision-language alignment, while end layers exhibit static sink behavior.
    \item We propose SAGE as a controllable local ROI-guided intervention tool that enables systematic layer sweeps and provides evidence that selecting an effective decoder layer for intervention can improve grounding and benchmark performance across multiple VLMs.
\end{itemize}

\section{Related Work}
\label{sec:related_work}

\subsection{Vision-Language Models}
\label{sec:related_vlm}
Large vision-language models (VLMs) couple visual representation learning with language generation and reasoning. A dominant family follows the vision-encoder + decoder-only LLM design, in which image features are projected into visual tokens and jointly processed with text tokens by an autoregressive decoder. Representative examples include LLaVA-style models~\cite{liu2024improved,li2025llava} and Qwen-VL~\cite{bai2023qwen,wang2024qwen2,yang2025qwen3}, while recent work further improves efficiency and fine-grained perception by sparsifying visual tokens at inference or by predicting regions of interest to focus computation~\cite{khaki2025sparsevila,shi2026catching}. Other work explores alternative multimodal bridging and scaling strategies, including Flamingo~\cite{alayrac2022flamingo}, BLIP-2/InstructBLIP~\cite{li2023blip,dai2023instructblip}, explicit grounding formulations~\cite{geigle2024does}, and multilingual large-scale training as in PaLI and PaLI-X~\cite{chen2022pali,chen2023pali}.

\subsection{Attention Sinks and Visual Attention Sinks}
\label{sec:related_sink}

Transformer attention is known to exhibit attention sinks, where a few tokens absorb disproportionate attention mass independent of semantic relevance. This phenomenon has been linked to efficient long-context and streaming inference, and later work studies when sinks emerge during pretraining and how architectural or normalization choices affect their formation~\cite{xiao2023efficient,gu2025attention,wu2024role}. Similar pathologies also arise over visual tokens in multimodal models. VAR~\cite{kang2025see} extends the sink notion to LVLMs and shows repeated concentration on query-irrelevant visual tokens; related work in vision transformers and token/head-level analyses likewise studies architectural mitigation, selective steering toward salient regions, and training-time regularization for stronger grounding~\cite{feng2026edit,tang2026not,zhao2026tell,esmaeilkhani2026direct}.

These attention pathologies are closely connected to hallucination and grounding failures, where outputs are not faithfully supported by the image. Benchmarks such as POPE~\cite{li2023evaluating} quantify object hallucination, while recent methods mitigate it through decoding-time interventions, contrastive signals, or vision-side masking of uncertain visual tokens~\cite{kim2024code,park2025convis,zhao2026tell,chen2025valse,jung2025visualattention,seo2026epistemic,leng2024mitigating}.

\noindent\textbf{Our positioning and new finding.}
While prior work establishes the existence of visual attention sinks, our study shows that sink behavior is structured across depth rather than uniform. Early and late decoder layers tend to exhibit prompt-invariant collapse onto stable image regions, whereas mid layers remain comparatively prompt-responsive and play the main role in vision-language alignment. This layer-dependent split motivates interventions that steer attention away from prompt-invariant sink regions and query-dependent ROI.


\section{Methodology}
\label{sec:methodology}

In this section, we present \textbf{SAGE} (Sink-Aware Guided Emphasis), a training-free inference-time intervention that steers decoder attention toward query-relevant visual evidence and away from attention-sink regions.

We focus on vision-encoder + decoder-only LLM VLMs, including LLaVA 1.5\cite{liu2024improved}, Qwen2-VL~\cite{wang2024qwen2}, InternVL3~\cite{zhu2025internvl3}, and NVILA~\cite{liu2025nvila}, where a vision encoder produces visual features that are projected or adapted into a sequence of visual tokens and processed jointly with text tokens by an autoregressive LLM via causal self-attention.

A key of SAGE is that it can be instantiated with multiple backbone-derived ROI generators as long as they can be mapped into a token-aligned ROI mask over visual tokens.
In this paper, we instantiate SAGE with ROI masks derived from standard vision backbones such as CLIP~\cite{radford2021learning}, ViT~\cite{dosovitskiy2020image}, and DINOv3~\cite{simeoni2025dinov3}.
These masks provide a simple and consistent interface for decoder-side attention steering across models.

\begin{figure*}[t]
    \centering
    \includegraphics[width=0.85\textwidth]{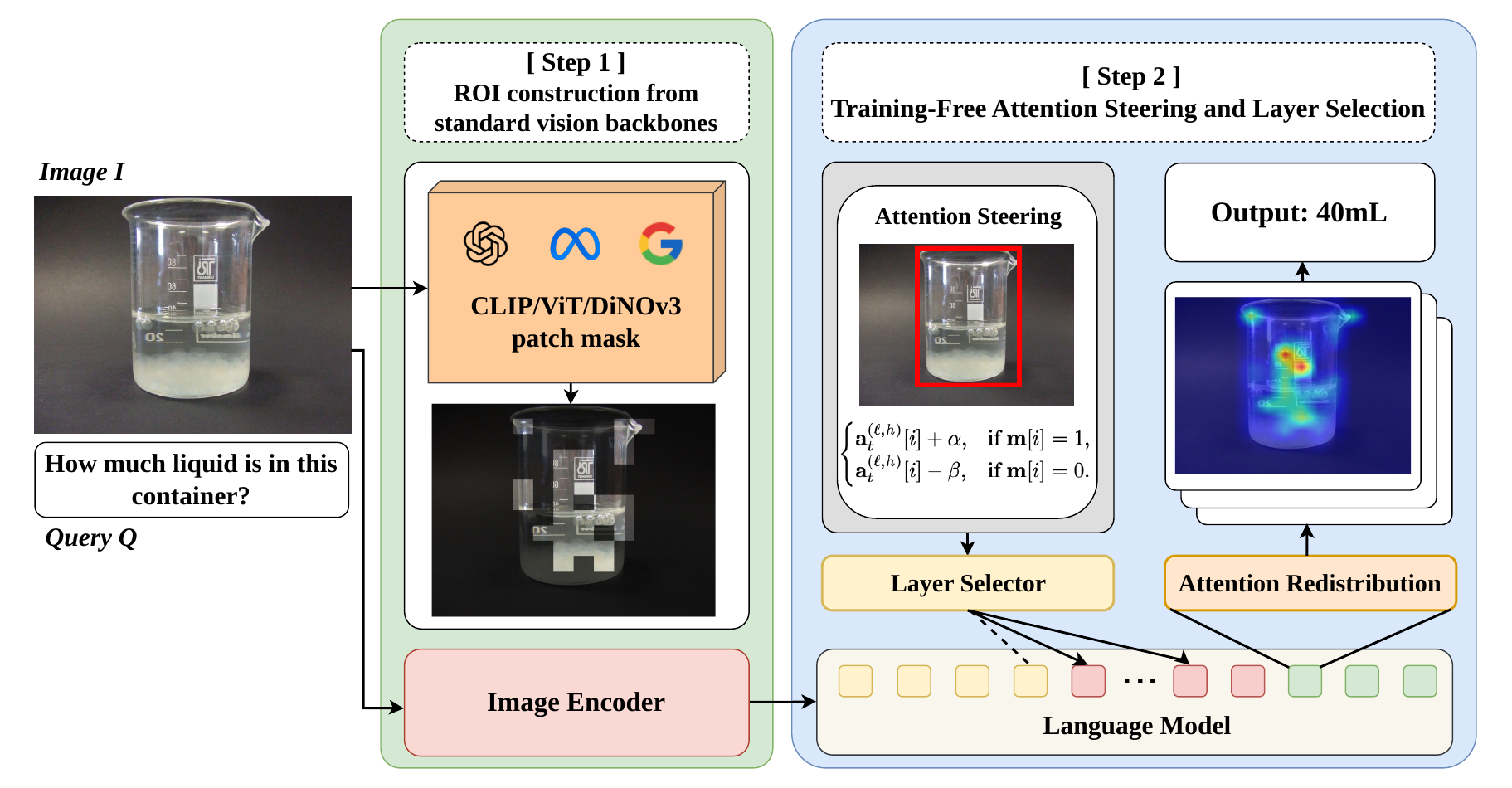}
    \caption{\textbf{Overview of SAGE.}
    Given an image $I$ and a query $Q$, SAGE operates in two stages.
    \textbf{Step 1 (ROI construction from standard vision backbones):} a token-aligned ROI mask over visual tokens is derived from patch-level relevance maps obtained from CLIP, ViT, or DINOv3.
    \textbf{Step 2 (Training-free attention steering and layer selection):} using the ROI mask, SAGE applies an additive bias to visual-token attention logits (Eq.~\eqref{eq:sage_logit_intervention}) at selected decoder layer(s), redistributing attention away from sink regions and toward query-relevant visual evidence.
    Layer selection can be performed via a layer sweep and top-$k$ selection strategy (Section~\ref{sec:method_layersel}).
    The resulting attention redistribution improves visual grounding during autoregressive generation.}
    \label{fig:architecture}
    \vspace{-4mm}
\end{figure*}

\subsection{Preliminaries and Notation}
\label{sec:method_setup}

Given an image $I$ and a query $Q$, the vision encoder produces visual features that are transformed into $N$ visual tokens
$\mathbf{V}=[v_1,\dots,v_N]$ via a projector/adapter.
The LLM based on the decoder-only processes a unified multimodal sequence consisting of visual tokens and text tokens and generates an answer autoregressively.

At decoding step $t$, layer $\ell\in\{0,\dots,L-1\}$, and head $h$, we denote by
$\mathbf{a}^{(\ell,h)}_t \in \mathbb{R}^{N}$
the restricted pre-softmax attention logits from the query token at step $t$ to the $N$ visual-token keys.
SAGE represents the guided ROI in visual-token space as a binary mask $\mathbf{m}\in{0,1}^{N}$, where $\mathbf{m}[i]=1$ marks the $i$-th token as inside the ROI.

\subsection{ROI Construction from Standard Vision Backbones}
\label{sec:method_roi}

SAGE requires only a token-aligned binary ROI mask $\mathbf{m}\in\{0,1\}^{N}$
and makes no assumption about how it is produced. We instantiate it with three
standard pretrained vision backbones, which differ in an important respect:
only CLIP conditions on the query, while ViT and DINOv3 produce query-agnostic saliency.

\noindent\textbf{CLIP} We project and $L_2$-normalize the
patch-token embeddings and the query text embedding into the shared CLIP space
and score each patch by $\cos(\text{patch},\text{text})$. The resulting
relevance map therefore depends on the query.

\noindent\textbf{ViT} We use attention rollout from the CLS token at the
last layer, averaged over heads. The query is not an input.

\noindent\textbf{DINOv3} We use CLS-to-patch attention at the last layer,
excluding register tokens. The query is not an input.

\noindent\textbf{Patch-to-token alignment.} For every source, the patch score map is
resized to the visual-token grid of the evaluated VLM with nearest-neighbor
interpolation, and the top-$k$ tokens are kept as the binary mask. 

\begin{figure*}[t]
    \centering
    \includegraphics[width=\textwidth]{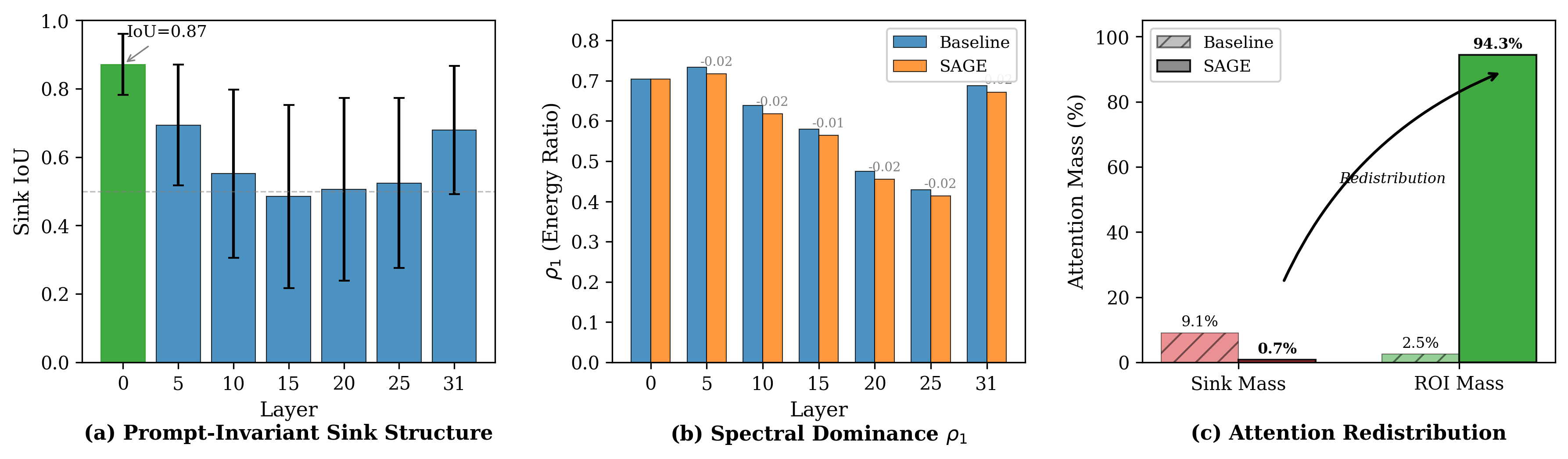}
    \caption{\textbf{Evidence for sink structure, spectral dominance, and attention redistribution.}
    (a) \textbf{Prompt-invariant sink structure.} For each layer, we form a sink set as the top-$5\%$ visual tokens by attention weight and report the IoU of sink sets across different prompts for the same image (error bars: standard deviation).
    (b) \textbf{Spectral dominance.} We stack visual-key attention logits across prompts and decoding steps to form $\mathbf{A}^{(\ell,h)}$ (Eq.~\eqref{eq:stacked_logits}) and report the top-1 energy ratio $\rho_1$ (Eq.~\eqref{eq:energy_ratio}); SAGE reduces spectral concentration, indicating weaker low-rank sink dominance.
    (c) \textbf{Attention redistribution.} we report attention mass assigned to sink tokens (top-$5\%$ tokens) and ROI tokens; SAGE shifts attention budget away from sink tokens and toward ROI, consistent with the odds-ratio effect in Eq.~\eqref{eq:odds_ratio}.}
    \label{fig:mechanism}
    \vspace{-4mm}
\end{figure*}

\subsection{Training-Free Attention Steering}
\label{sec:method_intervention}

Given a token-aligned ROI mask $\mathbf{m}\in\{0,1\}^{N}$, SAGE steers attention toward query-relevant visual evidence by directly biasing attention allocation within the decoder-only LLM.
For a chosen layer $\ell$ and head $h$ at decoding step $t$, let $\mathbf{a}^{(\ell,h)}_t\in\mathbb{R}^{N}$ denote the (visual-token) attention logits over $N$ visual tokens.
SAGE applies a simple additive bias controlled by $\alpha>0$ (ROI amplification) and $\beta>0$ (non-ROI suppression):
\begin{equation}
\mathbf{a}'^{(\ell,h)}_t[i] =
\begin{cases}
\mathbf{a}^{(\ell,h)}_t[i] + \alpha, & \text{if } \mathbf{m}[i] = 1, \\
\mathbf{a}^{(\ell,h)}_t[i] - \beta,  & \text{if } \mathbf{m}[i] = 0.
\end{cases}
\label{eq:sage_logit_intervention}
\end{equation}
This bias reallocates attention mass toward the guided ROI, encouraging the model to consult query-relevant visual tokens while preserving the original model parameters.
SAGE is therefore training-free and can be applied at inference time to arbitrary layer sets $\mathcal{L}_{\mathrm{int}}$ (single-layer or multi-layer).

\noindent\textbf{Implementation note.}
The ROI mask is first computed at patch level and then aligned to the visual-token grid used by each evaluated VLM.
The exact patch-to-token mapping depends on the model's vision encoder and preprocessing pipeline; we keep this mapping fixed within each experiment and summarize the details in Appendix~\ref{app:roi_details}.

\subsection{Layer Sweep and Top-$k$ Layer Selection}
\label{sec:method_layersel}

SAGE can be applied to different decoder depths, and its effect is generally layer-dependent.
Rather than prescribing a single fixed layer as the only valid choice, we use a simple sweep protocol to (i) characterize where attention steering tends to be effective and (ii) form reasonable intervention sets (e.g., top-$k$ layers) that perform well in practice.

Concretely, we run a single-layer sweep where SAGE is applied to one layer $\ell$ at a time (all other layers unchanged) and measure the relative change in performance against the no-intervention baseline.
Let $\Delta(\ell)$ denote an aggregated relative performance change for intervening only at layer $\ell$.
This sweep serves as a diagnostic: it highlights layers where steering is more effective and provides a ranking that can be used to construct multi-layer intervention sets.
Using the ranking by $\Delta(\ell)$, we define a top-$k$ layers set
\begin{equation}
\mathcal{L}_{\mathrm{top}\text{-}k} = \mathrm{TopK}_{\ell}\big(\Delta(\ell)\big),
\end{equation}
and apply SAGE to $\mathcal{L}_{\mathrm{top}\text{-}k}$ in subsequent experiments as one practical choice.
Importantly, our goal is not to claim that only a particular layer (or only one selection rule) works; instead, the sweep illustrates that attention steering itself is broadly effective, and that multiple reasonable layer selections (single-layer or multi-layer) can yield improvements.

\subsection{Mechanistic Rationale}
\label{sec:method_theory}

We provide a mechanistic view of why SAGE can improve visual grounding in terms of Sink Subspace and Attention Redistribution.
Our analysis in Figure~\ref{fig:mechanism} suggests that visual attention sinks in VLM decoders are not arbitrary noise but exhibit structured, prompt-invariant patterns across depth.
We formalize this intuition using a simple spectral perspective and show how the SAGE bias induces an attention-redistribution effect.
Figure~\ref{fig:mechanism} summarizes the key empirical signatures discussed below.

\noindent\textbf{Sink subspace perspective.}
Fix a decoder layer $\ell$ and head $h$.
For each decoding step $t$ and prompt instance $p$, let $\mathbf{a}^{(\ell,h)}_{p,t}\in\mathbb{R}^{N}$ denote the pre-softmax attention logits over the $N$ visual-token keys (as in Section~\ref{sec:method_setup}).
To quantify prompt-invariant sink, we define a sink set as the top-$r\%$ visual tokens by attention weight and measure its overlap across prompts for the same image using set IoU (Fig.~\ref{fig:mechanism}a).

We stack these logit vectors across prompts and decoding steps into a matrix
\begin{equation}
\mathbf{A}^{(\ell,h)} \in \mathbb{R}^{M \times N}, 
\qquad
\mathbf{A}^{(\ell,h)}[r,:] = \mathbf{a}^{(\ell,h)}_{p,t},
\label{eq:stacked_logits}
\end{equation}
where $r$ indexes the pair $(p,t)$ and $M$ is the total number of stacked instances.
We then compute an SVD,
\begin{equation}
\mathbf{A}^{(\ell,h)} = \mathbf{U}\mathbf{\Sigma}\mathbf{V}^{\top}.
\label{eq:svd}
\end{equation}
If attention repeatedly collapses onto a small set of prompt-invariant sink tokens, the spectrum becomes more concentrated, and the top right-singular vectors (columns of $\mathbf{V}$) capture stable token-level directions corresponding to sink-dominated patterns.
A convenient summary is the energy captured by the top-$k$ components:
\begin{equation}
\rho_k(\mathbf{A}) = \frac{\sum_{j=1}^{k}\sigma_j^2}{\sum_{j=1}^{\min(M,N)}\sigma_j^2},
\label{eq:energy_ratio}
\end{equation}
where $\{\sigma_j\}$ are singular values.
Our hypothesis is that prompt-invariant sinks induce high $\rho_k$ for small $k$, reflecting structured low-rank dominance (Fig.~\ref{fig:mechanism}b).

\noindent\textbf{Attention redistribution induced by SAGE.}
SAGE applies an additive logit bias to visual-token attention as defined in Eq.~\eqref{eq:sage_logit_intervention}.
Let $p_i=\mathrm{Softmax}(\mathbf{a})_i$ and $p'_i=\mathrm{Softmax}(\mathbf{a}')_i$ be the attention weights over visual keys before and after applying Eq.~\eqref{eq:sage_logit_intervention}, respectively.
For any ROI index $i$ and non-ROI index $j$, the odds ratio satisfies
\begin{equation}
\frac{p'_i}{p'_j} = \exp(\alpha+\beta)\cdot \frac{p_i}{p_j}.
\label{eq:odds_ratio}
\end{equation}
Consequently, even when sink locations remain spatially stable, suppressing non-ROI tokens (which may include sink tokens) and amplifying ROI tokens can substantially redistribute attention budget toward query-relevant visual evidence in our analysis setting (Fig.~\ref{fig:mechanism}c).

\noindent\textbf{Testable implications.}
This view suggests two measurable signatures of effective attention steering:
(i) reduced sink dominance reflected by lower spectral concentration (e.g., smaller $\rho_k$ for small $k$), and
(ii) increased attention mass on ROI tokens accompanied by decreased mass on sink tokens.
We examine these signatures with layer-wise spectral and mass-based analyses, with details provided in Appendix~\ref{app:mech}.

\noindent\textbf{Cross-model evidence.} The two diagnostics are not independent
descriptions of the same plot. Across LLaVA-1.5 and Qwen2-VL on
TextVQA and GQA, layer-wise prompt-invariant Sink IoU correlates strongly with
top-1 spectral concentration (Pearson $r=0.886$--$0.942$). This supports
reading prompt-invariant sinks as structured low-rank routing directions
rather than image-specific heatmap artifacts, and it is why the two can be
combined into a single label-free depth criterion
(Section~\ref{sec:method_layersel}).

\begin{table*}[hbt!]
\centering
\vspace{-2mm}
\caption{\textbf{Main results of SAGE across VLM families.}
For each model, benchmark, and ROI source, the reported SAGE score corresponds to the best single-layer result obtained from the layer sweep.
Exact selected layer indices are listed in Appendix Table~\ref{tab:best_layers_main}.}
\vspace{-2mm}
\label{tab:main_results}
\begin{small}
\begin{sc}
\renewcommand{\arraystretch}{1.1}

\definecolor{posblue}{RGB}{0, 102, 204}
\definecolor{negred}{RGB}{204, 0, 0}

\resizebox{\linewidth}{!}
{
\begin{tabular}{l|ccccccccccc}
\toprule
\multirow{2}{*}{Model} & \multirow{2}{*}{TextVQA} & \multicolumn{2}{c}{POPE} & \multirow{2}{*}{GQA} & \multirow{2}{*}{ScienceQA} & \multirow{2}{*}{MME} & \multicolumn{2}{c}{MMVP} & \multicolumn{2}{c}{MMVP Original} & \multirow{2}{*}{MMVet} \\
\cmidrule(lr){3-4} \cmidrule(lr){8-9} \cmidrule(lr){10-11}
& & Acc & F1 & & & & Image & Pair & Image & Pair & \\
\midrule
LLaVA 1.5-13B & 55.50 & 82.75 & 81.55 & 61.65 & 63.98 & 1783.11 & 61.48 & 11.85 & 33.58 & 10.65 & 31.20 \\
\hline
& \textbf{61.94} & \textbf{86.84} & \textbf{86.58} & 62.65 & 68.63 & 1842.58 & 62.96 & 37.78 & 50.00 & \textbf{22.00} & \textbf{42.20} \\
\multirow{-2}{*}{+ SAGE (CLIP)} & \scriptsize(\textcolor{posblue}{+6.44}) & \scriptsize(\textcolor{posblue}{+4.09}) & \scriptsize(\textcolor{posblue}{+5.03}) & \scriptsize(\textcolor{posblue}{+1.00}) & \scriptsize(\textcolor{posblue}{+4.65}) & \scriptsize(\textcolor{posblue}{+59.47}) & \scriptsize(\textcolor{posblue}{+1.48}) & \scriptsize(\textcolor{posblue}{+25.93}) & \scriptsize(\textcolor{posblue}{+16.42}) & \scriptsize(\textcolor{posblue}{+11.35}) & \scriptsize(\textcolor{posblue}{+11.00}) \\
\hline
& 61.85 & 85.81 & 84.82 & \textbf{65.50} & 69.15 & \textbf{1895.55} & \textbf{68.89} & \textbf{38.52} & \textbf{58.58} & 21.27 & 34.50 \\
\multirow{-2}{*}{+ SAGE (ViT)} & \scriptsize(\textcolor{posblue}{+6.35}) & \scriptsize(\textcolor{posblue}{+3.06}) & \scriptsize(\textcolor{posblue}{+3.27}) & \scriptsize(\textcolor{posblue}{+3.85}) & \scriptsize(\textcolor{posblue}{+5.17}) & \scriptsize(\textcolor{posblue}{+112.44}) & \scriptsize(\textcolor{posblue}{+7.41}) & \scriptsize(\textcolor{posblue}{+26.67}) & \scriptsize(\textcolor{posblue}{+25.00}) & \scriptsize(\textcolor{posblue}{+10.62}) & \scriptsize(\textcolor{posblue}{+3.30}) \\
\hline
& 61.54 & 83.00 & 81.72 & 62.55 & \textbf{69.17} & 1791.44 & 63.33 & 14.07 & 34.00 & 12.00 & 36.47 \\
\multirow{-2}{*}{+ SAGE (DINOv3)} & \scriptsize(\textcolor{posblue}{+6.04}) & \scriptsize(\textcolor{posblue}{+0.25}) & \scriptsize(\textcolor{posblue}{+0.17}) & \scriptsize(\textcolor{posblue}{+0.90}) & \scriptsize(\textcolor{posblue}{+5.19}) & \scriptsize(\textcolor{posblue}{+8.33}) & \scriptsize(\textcolor{posblue}{+1.85}) & \scriptsize(\textcolor{posblue}{+2.22}) & \scriptsize(\textcolor{posblue}{+0.42}) & \scriptsize(\textcolor{posblue}{+1.35}) & \scriptsize(\textcolor{posblue}{+5.27}) \\
\midrule

Qwen2-VL-7B & 81.32 & 75.94 & 75.42 & 62.04 & 68.34 & 1792.60 & 78.15 & 57.78 & 69.33 & 40.67 & 59.60 \\
\hline
& 81.43 & 76.59 & 75.52 & \textbf{62.38} & \textbf{74.16} & 1895.10 & 81.15 & 58.28 & \textbf{71.21} & 41.33 & 61.10 \\
\multirow{-2}{*}{+ SAGE (CLIP)} &
\scriptsize(\textcolor{posblue}{+0.11}) & \scriptsize(\textcolor{posblue}{+0.65}) & \scriptsize(\textcolor{posblue}{+0.10}) & \scriptsize(\textcolor{posblue}{+0.34}) &
\scriptsize(\textcolor{posblue}{+5.82}) &
\scriptsize(\textcolor{posblue}{+102.50}) &
\scriptsize(\textcolor{posblue}{+3.00}) &
\scriptsize(\textcolor{posblue}{+0.50}) &
\scriptsize(\textcolor{posblue}{+1.88}) &
\scriptsize(\textcolor{posblue}{+0.66}) &
\scriptsize(\textcolor{posblue}{+1.50}) \\
\hline
& \textbf{81.52} & 77.15 & 76.81 & 62.28 & 71.28 & \textbf{1920.00} & 78.88 & 57.94 & 70.24 & \textbf{42.53} & \textbf{61.80} \\
\multirow{-2}{*}{+ SAGE (ViT)} &
\scriptsize(\textcolor{posblue}{+0.20}) &
\scriptsize(\textcolor{posblue}{+1.21}) & \scriptsize(\textcolor{posblue}{+1.39}) &
\scriptsize(\textcolor{posblue}{+0.24}) &
\scriptsize(\textcolor{posblue}{+2.94}) &
\scriptsize(\textcolor{posblue}{+127.40}) &
\scriptsize(\textcolor{posblue}{+0.73}) &
\scriptsize(\textcolor{posblue}{+0.16}) &
\scriptsize(\textcolor{posblue}{+0.91}) &
\scriptsize(\textcolor{posblue}{+1.86}) &
\scriptsize(\textcolor{posblue}{+2.20}) \\
\hline
& 81.41 & \textbf{78.35} & \textbf{77.56} & 62.15 & 68.47 & 1795.23 & \textbf{91.85} & \textbf{85.19} & 70.67 & \textbf{42.53} & 59.90 \\
\multirow{-2}{*}{+ SAGE (DINOv3)} &
\scriptsize(\textcolor{posblue}{+0.09}) & \scriptsize(\textcolor{posblue}{+2.41}) & \scriptsize(\textcolor{posblue}{+2.14}) &
\scriptsize(\textcolor{posblue}{+0.11}) &
\scriptsize(\textcolor{posblue}{+0.13}) &
\scriptsize(\textcolor{posblue}{+2.63}) &
\scriptsize(\textcolor{posblue}{+13.70}) &
\scriptsize(\textcolor{posblue}{+27.41}) &
\scriptsize(\textcolor{posblue}{+1.34}) &
\scriptsize(\textcolor{posblue}{+1.86}) &
\scriptsize(\textcolor{posblue}{+0.30}) \\
\midrule

InternVL3-2B & 47.60 & 87.97 & 87.41 & 56.85 & 61.95 & 2087.90 & 73.44 & 49.37 & 72.67 & 47.33 & 61.40 \\
\hline
& 60.20 & 88.27 & 88.36 & 57.15 & 62.25 & 2092.58 & 76.30 & \textbf{58.52} & 73.33 & 49.33 & \textbf{62.20} \\
\multirow{-2}{*}{+ SAGE (CLIP)} &
\scriptsize(\textcolor{posblue}{+12.60}) &
\scriptsize(\textcolor{posblue}{+0.30}) & \scriptsize(\textcolor{posblue}{+0.95}) & \scriptsize(\textcolor{posblue}{+0.30}) & \scriptsize(\textcolor{posblue}{+0.30}) &
\scriptsize(\textcolor{posblue}{+4.68}) & \scriptsize(\textcolor{posblue}{+2.86}) & \scriptsize(\textcolor{posblue}{+9.15}) & \scriptsize(\textcolor{posblue}{+0.66}) & \scriptsize(\textcolor{posblue}{+2.00}) & \scriptsize(\textcolor{posblue}{+0.80}) \\
\hline
& 47.80 & 88.48 & 87.89 & \textbf{59.85} & \textbf{62.45} & \textbf{2097.29} & \textbf{77.78} & 57.78 & 72.33 & 46.00 & 62.10 \\
\multirow{-2}{*}{+ SAGE (ViT)} &
\scriptsize(\textcolor{posblue}{+0.20}) &
\scriptsize(\textcolor{posblue}{+0.51}) &
\scriptsize(\textcolor{posblue}{+0.48}) &
\scriptsize(\textcolor{posblue}{+3.00}) &
\scriptsize(\textcolor{posblue}{+0.50}) &
\scriptsize(\textcolor{posblue}{+9.39}) & \scriptsize(\textcolor{posblue}{+4.34}) & \scriptsize(\textcolor{posblue}{+8.41}) & \scriptsize(\textcolor{negred}{-0.34}) &
\scriptsize(\textcolor{negred}{-1.33}) &
\scriptsize(\textcolor{posblue}{+0.70}) \\
\hline
& \textbf{73.50} & \textbf{88.89} & \textbf{88.60} & 57.55 & 62.02 & 2072.23 & 73.70 & 49.63 & \textbf{75.00} & \textbf{50.70} & 61.50 \\
\multirow{-2}{*}{+ SAGE (DINOv3)} &
\scriptsize(\textcolor{posblue}{+25.90}) &
\scriptsize(\textcolor{posblue}{+0.92}) &
\scriptsize(\textcolor{posblue}{+1.19}) &
\scriptsize(\textcolor{posblue}{+0.70}) &
\scriptsize(\textcolor{posblue}{+0.07}) &
\scriptsize(\textcolor{negred}{-15.67}) &
\scriptsize(\textcolor{posblue}{+0.26}) &
\scriptsize(\textcolor{posblue}{+0.26}) &
\scriptsize(\textcolor{posblue}{+2.33}) &
\scriptsize(\textcolor{posblue}{+3.37}) &
\scriptsize(\textcolor{posblue}{+0.10}) \\
\midrule

NVILA-8B & 74.44 & 85.37 & 84.81 & 62.20 & 87.57 & 2107.29 & 78.52 & 58.52 & 70.67 & 43.33 & 61.79 \\
\hline
& \textbf{78.13} & \textbf{92.20} & \textbf{91.99} & 62.80 & 89.37 & 2125.59 & 78.88 & 59.25 & \textbf{72.67} & \textbf{47.33} & 62.16 \\
\multirow{-2}{*}{+ SAGE (CLIP)} &
\scriptsize(\textcolor{posblue}{+3.69}) &
\scriptsize(\textcolor{posblue}{+6.83}) &
\scriptsize(\textcolor{posblue}{+7.18}) &
\scriptsize(\textcolor{posblue}{+0.60}) &
\scriptsize(\textcolor{posblue}{+1.80}) &
\scriptsize(\textcolor{posblue}{+18.30}) &
\scriptsize(\textcolor{posblue}{+0.36}) &
\scriptsize(\textcolor{posblue}{+0.73}) &
\scriptsize(\textcolor{posblue}{+2.00}) &
\scriptsize(\textcolor{posblue}{+4.00}) &
\scriptsize(\textcolor{posblue}{+0.37}) \\
\hline
& 74.84 & 88.95 & 88.26 & 64.90 & \textbf{92.24} & \textbf{2133.59} & 79.12 & 59.82 & 71.48 & 45.13 & 61.94 \\
\multirow{-2}{*}{+ SAGE (ViT)} &
\scriptsize(\textcolor{posblue}{+0.40}) &
\scriptsize(\textcolor{posblue}{+3.58}) &
\scriptsize(\textcolor{posblue}{+3.45}) &
\scriptsize(\textcolor{posblue}{+2.70}) &
\scriptsize(\textcolor{posblue}{+4.67}) &
\scriptsize(\textcolor{posblue}{+26.30}) &
\scriptsize(\textcolor{posblue}{+0.60}) &
\scriptsize(\textcolor{posblue}{+1.30}) &
\scriptsize(\textcolor{posblue}{+0.81}) &
\scriptsize(\textcolor{posblue}{+1.80}) &
\scriptsize(\textcolor{posblue}{+0.15}) \\
\hline
& 74.75 & 89.00 & 88.18 & \textbf{64.95} & 91.61 & 2118.85 & \textbf{79.94} & \textbf{60.02} & 71.86 & 44.94 & \textbf{62.36} \\
\multirow{-2}{*}{+ SAGE (DINOv3)} &
\scriptsize(\textcolor{posblue}{+0.31}) & \scriptsize(\textcolor{posblue}{+3.63}) & \scriptsize(\textcolor{posblue}{+3.37}) &
\scriptsize(\textcolor{posblue}{+2.75}) &
\scriptsize(\textcolor{posblue}{+4.04}) &
\scriptsize(\textcolor{posblue}{+11.56}) &
\scriptsize(\textcolor{posblue}{+1.42}) &
\scriptsize(\textcolor{posblue}{+1.50}) &
\scriptsize(\textcolor{posblue}{+1.19}) &
\scriptsize(\textcolor{posblue}{+1.61}) &
\scriptsize(\textcolor{posblue}{+0.57}) \\
\bottomrule

\end{tabular}
}
\vspace{-4mm}
\end{sc}
\end{small}
\end{table*}


\section{Experiments}
\label{sec:experiments}

\subsection{Experimental Setup}
\label{sec:exp_setup}

\noindent\textbf{Models.}
We evaluate SAGE on vision-encoder + decoder-only LLM VLMs from the LLaVA 1.5~\cite{liu2024improved}, Qwen2-VL~\cite{wang2024qwen2}, InternVL3~\cite{zhu2025internvl3} and NVILA~\cite{liu2025nvila}.
Unless stated otherwise, we use the instruct-tuned checkpoints listed in Table~\ref{tab:main_results}.

\noindent\textbf{ROI sources.}
Our main results use ROI masks derived from standard vision backbones, specifically CLIP, ViT, and DINOv3.
Given an image and query, each source produces patch-level relevance scores, which are thresholded to obtain a binary ROI mask aligned to the visual-token grid of the evaluated VLM.
Detailed source-specific settings are summarized in Appendix~\ref{app:roi_sources}.

\noindent\textbf{Intervention settings.}
Unless stated otherwise, we use a fixed bias strength across models.
In all experiments, we set $\alpha=5.0$ (ROI amplification) and $\beta=2.0$ (non-ROI bias).
For layer-wise ablations, we apply a single-layer intervention by sweeping the intervention layer across the decoder stack of each model, with all other layers unchanged.


\noindent\textbf{Normalized decoder depth.}
Because different VLM families can have different numbers of decoder layers, we use normalized decoder depth when comparing layer-wise behavior across models.
For a model with $L$ decoder layers and zero-indexed layer $\ell$, we define the normalized depth as $d(\ell)=\ell/(L-1)$.
We use raw layer indices only in model-specific analyses such as LLaVA-1.5 case studies and appendix tables.

\subsection{Main Evaluations}
\label{sec:exp_main}

We evaluate SAGE on representative vision-encoder + decoder-only LLM VLM families and summarize the main results in Table~\ref{tab:main_results}.
For each model, benchmark, and ROI source, the reported SAGE score corresponds to the best single-layer result obtained from the layer sweep.
We use these results to quantify the peak effect of SAGE at effective decoder depths, rather than to claim a universal fixed layer rule.
Because the layer is selected per benchmark, these values indicate an upper
bound on the achievable effect. Section~\ref{sec:exp_controls} reports the result
obtained with a single layer fixed in advance, without benchmark labels.
For transparency, the exact selected layer indices are listed in Appendix Table~\ref{tab:best_layers_main}; when discussing layer behavior across models, we refer to normalized decoder depth rather than raw layer number.

Across the evaluated VLMs, SAGE improves over the no-intervention baseline on many benchmarks and across multiple ROI sources.
These gains are consistent with our motivating hypothesis that decoder-side attention pathologies can suppress query-relevant visual evidence, and that a lightweight logit bias toward an ROI helps the model focus on more relevant visual tokens.
Across CLIP, ViT, and DINOv3, the qualitative trend is similar: the strongest improvements appear at selected decoder depths, while the effective depth can vary across model families and benchmarks.

\begin{figure}[t]
\centering
\includegraphics[width=\linewidth]{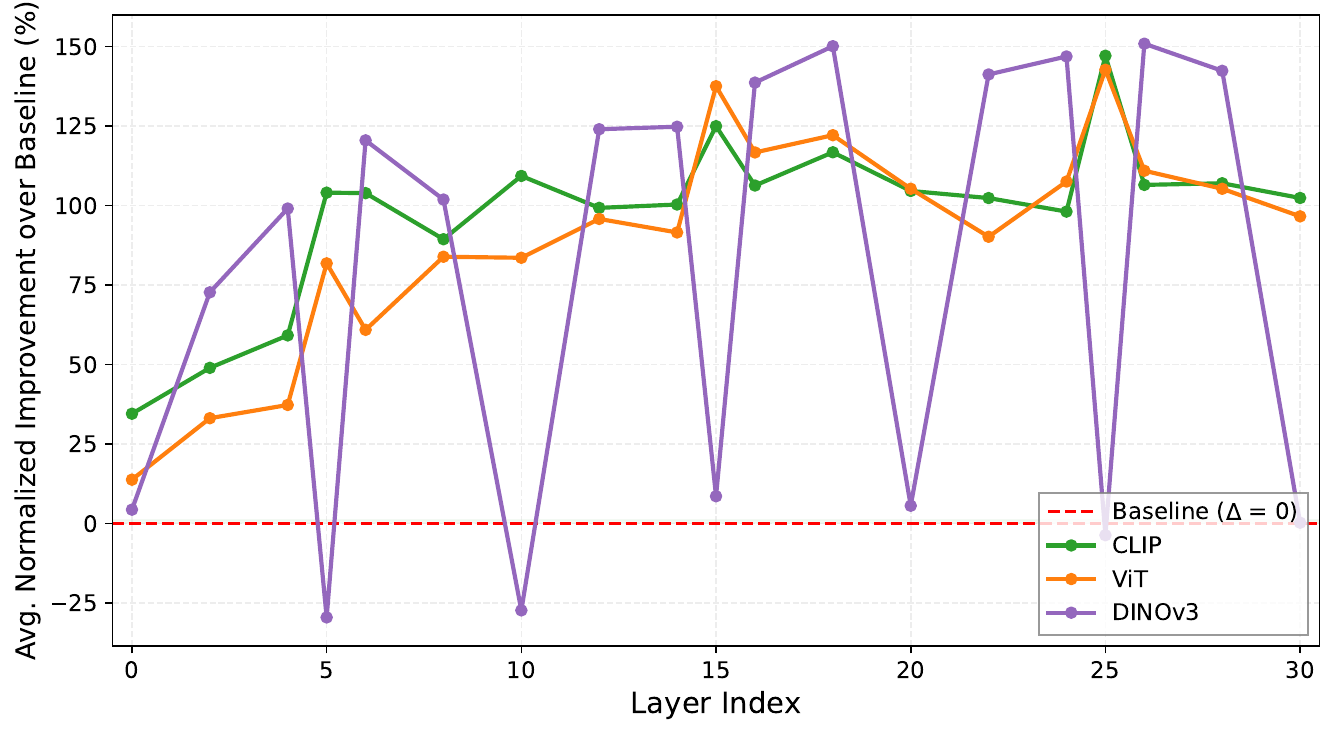}
\vspace{-1em}
\caption{\textbf{Layer-wise averaged normalized improvement on MMVP-family benchmarks.}
For each ROI source (CLIP, ViT, DINOv3), we apply SAGE to one decoder layer and evaluate on four fine-grained sub-benchmarks: MMVP image, MMVP pair, MMVP-Original image, and MMVP-Original pair.
For each sub-benchmark, we compute the normalized improvement over the no-intervention baseline as $(s-b)/b \times 100$, where $s$ is the score with SAGE and $b$ is the baseline score.
The plotted value is the average of these four normalized improvements at each intervention layer.
The x-axis shows the LLaVA-1.5 decoder layer index, and values above zero indicate improvement over the baseline.}
\label{fig:mmvp_avg_delta}
\vspace{-1em}
\end{figure}

\noindent\textbf{Fine-grained visual discrimination.}
The MMVP-family benchmarks are particularly relevant to our setting because they emphasize fine-grained pairwise visual discrimination, where the decisive evidence is often local and easy to miss.
Table~\ref{tab:main_results} shows strong improvements on MMVP and MMVP-Original, especially on pairwise metrics where both related samples must be answered correctly.
To further examine whether these gains are tied only to a single selected layer, Figure~\ref{fig:mmvp_avg_delta} reports layer-wise averaged normalized improvement on the four MMVP-family sub-benchmarks for LLaVA-1.5.
For CLIP-, ViT-, and DINOv3-derived ROI masks, the averaged improvement remains positive across a broad range of decoder layers, indicating that the benefit on fine-grained visual discrimination is not restricted to one isolated intervention depth.

\begin{figure}[t]
\centering
\includegraphics[width=\linewidth]{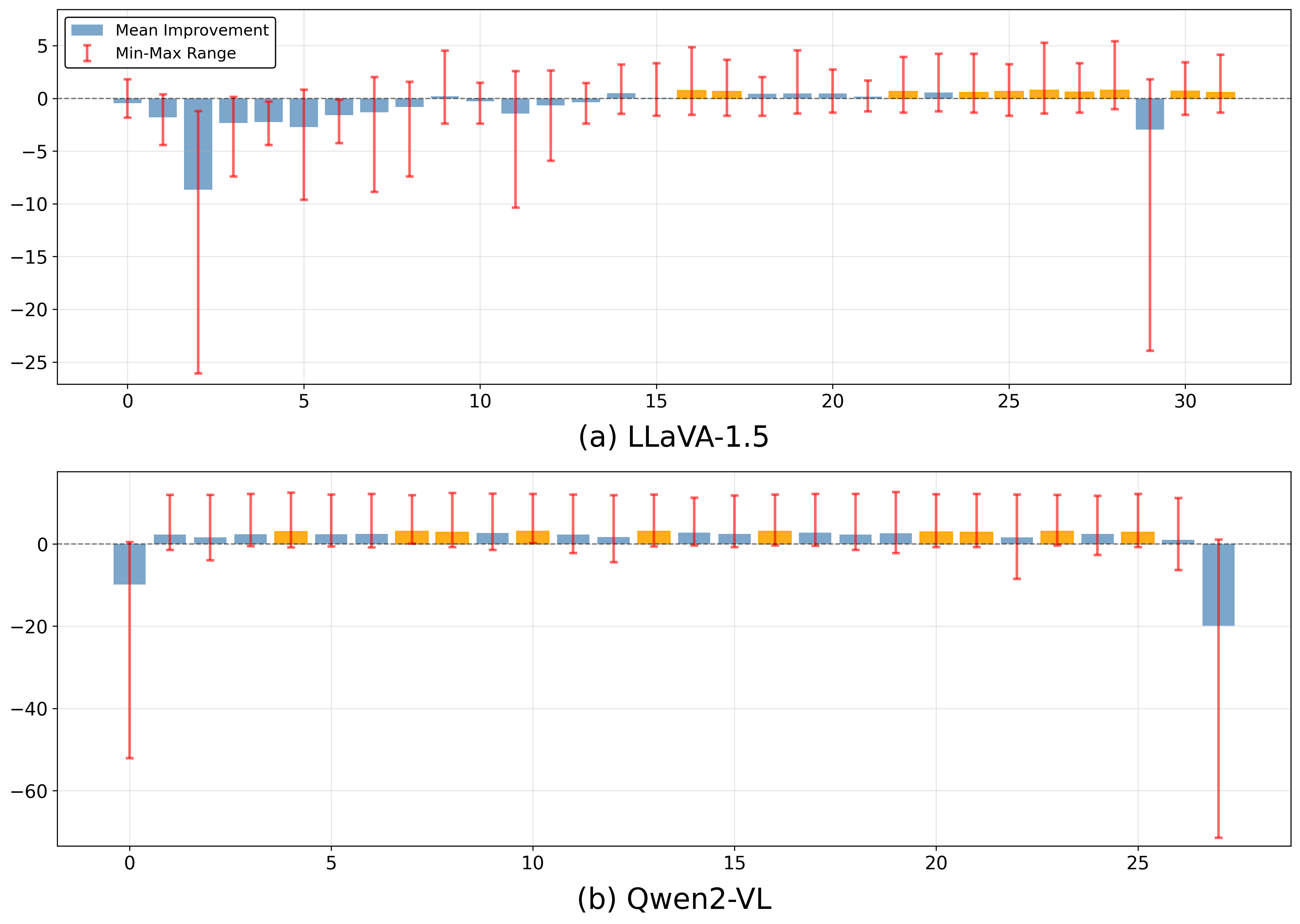} 
\vspace{-1em}
\caption{\textbf{Layer-wise single-layer sweep of SAGE interventions.}
We report relative performance improvement over the no-intervention baseline when applying SAGE to a single decoder layer at a time.
The x-axis is normalized decoder depth $d(\ell)=\ell/(L-1)$, allowing comparison across models with different decoder depths.
The horizontal reference at $0\%$ indicates the no-intervention baseline.
Bars aggregate results across all benchmarks evaluated in our experiments, and error bars show the min--max range.
Orange bars indicate the top-performing normalized depth positions.}
\label{fig:AGAI_llava_qwen}
\vspace{-1em}
\end{figure}

\subsection{Layer-wise Effects}
\label{sec:exp_layerwise}

This section analyzes how SAGE’s effectiveness varies with normalized decoder depth on LLaVA-1.5 and Qwen2-VL.
Although SAGE uses the same intervention rule (Eq.~\eqref{eq:sage_logit_intervention}) and the same ROI construction procedure within each experiment setting, its effect can vary substantially depending on where in the decoder stack the intervention is applied.
To isolate this factor, we conduct a single-layer sweep over the decoder stack for each model family by setting $\mathcal{L}_{\mathrm{int}}=\{\ell\}$ and intervening at one layer at a time.
Figure~\ref{fig:AGAI_llava_qwen} summarizes the layer-wise improvements over the no-intervention baseline as a function of normalized decoder depth, aggregated across benchmarks (median), and highlights the top-performing depths.

\noindent\textbf{Layer-wise variability.}
Figure~\ref{fig:AGAI_llava_qwen} shows that SAGE exhibits clear layer-wise variability.
Across both model families, intervening at a subset of layers yields consistent gains, whereas intervening at other layers leads to smaller changes and can occasionally be neutral.
This indicates that attention steering is not uniformly beneficial throughout the decoder and that layer selection is a key factor in determining the effectiveness of SAGE.

\noindent\textbf{Connection to visual attention sinks.}
Prior work reports that VLMs can suffer from visual attention sinks, where attention repeatedly collapses onto a small set of query-irrelevant visual tokens and limits the use of task-relevant visual evidence.
SAGE is designed to counteract this pathology by reallocating attention toward a query-dependent ROI.
The concentration of gains in selected decoder layers suggests that correcting attention allocation at those depths can improve grounding and downstream performance in the evaluated setting.

\begin{table}[!htbp]
\centering
\small
\caption{\textbf{Depth-grouped summaries of multi-layer SAGE sets on TextVQA.}
Each value reports the mean accuracy within a coarse depth-composition group.}
\label{tab:depth_group_summary}
\begin{tabular}{l c}
\toprule
\textbf{Depth composition} & \textbf{Mean Acc.} \\
\midrule
\multicolumn{2}{c}{\textit{2-layer sets}} \\
\midrule
Very-early only & 19.19 \\
Early-mixed & 49.85 \\
Mid--mid & 61.96 \\
Mid--final & \textbf{62.60} \\
Final--final & 62.42 \\
\midrule
\multicolumn{2}{c}{\textit{3-layer sets}} \\
\midrule
Very-early $\times 3$ & 2.35 \\
Early $\times 3$ & 55.39 \\
Early--early--mid & 59.10 \\
Early--mid--mid & 60.73 \\
Mid $\times 3$ & 62.11 \\
Mid--mid--final & 62.10 \\
Mid--final--final & 62.40 \\
Final $\times 3$ & 62.40 \\
\bottomrule
\end{tabular}
\end{table}

\subsection{Depth-Grouped Multi-layer Comparisons}
\label{sec:exp_topk}

We further examine multi-layer SAGE by grouping the tested sets according to coarse normalized-depth composition.
Table~\ref{tab:depth_group_summary} reports mean TextVQA accuracy within each group, while selected exact LLaVA-1.5 layer combinations are deferred to Appendix Tables~\ref{tab:2layer_examples},~\ref{tab:3layer_examples}, and~\ref{tab:failure_examples}.

The summary shows a clear depth-dependent pattern.
Very-early groups are substantially weaker, whereas mid--final and final--final groups are stronger and more stable.
This indicates that multi-layer gains are not tied to a single exact layer set, but appear more broadly among non-early depth compositions.
Using normalized-depth groups reduces reliance on any single selected layer.
The gap between very-early and non-early groups suggests that the earliest decoder depths are more fragile intervention sites.
Additional exact combinations and a layer-skip ablation are provided in Appendix~\ref{app:multilayer_more} and Appendix~\ref{app:skip_ablation}.

\subsection{Fixed-Layer Result and ROI Content Control}
\label{sec:exp_controls}

Table~\ref{tab:main_results} selects the intervention layer per benchmark.
We therefore also evaluate a fixed-layer setting in which the intervention depth is chosen in advance, without benchmark labels, using the two diagnostics already introduced in Section~\ref{sec:method_theory}: prompt-invariant Sink IoU (Fig.~\ref{fig:mechanism}a) and top-1 spectral concentration $\rho_1$ (Eq.~\eqref{eq:energy_ratio}, Fig.~\ref{fig:mechanism}b).
Both are computed from attention statistics alone and require no labels.
In the LLaVA-1.5 diagnostic sweep, layer $\mathrm{L}24$ attains the lowest value under both, indicating the weakest prompt-invariant sink structure among the examined depths.
We select it before running any downstream evaluation and keep it fixed across all benchmarks, never re-selecting per benchmark or per metric.

Table~\ref{tab:controls} reports this setting against the re-measured baseline under one unified protocol (LLaVA-1.5-7B).
With one fixed layer, SAGE raises TextVQA from $44.0$ to $54.6$ and GQA from $57.1$ to $62.3$, while the changes on POPE, ScienceQA, MME, and MMVP are small.
This is consistent with the premise that the intervention helps when the decisive evidence is localized.

\noindent\textbf{ROI content control.}
Within this same fixed-layer setting, we further ask whether the gain depends on what the ROI contains.
A logit bias of this strength reallocates attention regardless of where the mask points, so we hold the intervention layer, the bias strength, and the mask size fixed and replace the guided ROI with a randomly selected set of visual tokens.
This control leaves every benchmark at or below baseline.
The effect therefore requires the content of the target region and is not a generic consequence of perturbing the attention distribution.

\begin{table}[t]
\centering
\caption{\textbf{Fixed-layer result and ROI content control}. \textbf{Best} selects the layer on the benchmark;
\textbf{L24} uses one fixed layer chosen by label-free diagnostics.
\textbf{Rand.}\ replaces the guided ROI with a random token set of the same
size, at the same layer and bias strength.}
\small
\begin{tabular}{@{}lcccc@{}}
\toprule
 & & \multicolumn{2}{c}{SAGE} & \\
\cmidrule(lr){3-4}
Benchmark & Base & Best & L24 & Rand. \\
\midrule
TextVQA               & 44.0 & 55.4 & 54.6 & 43.1 \\
GQA                   & 57.1 & 62.8 & 62.3 & 57.0 \\
POPE$_{\text{acc}}$   & 72.7 & 73.6 & 73.1 & 72.1 \\
POPE$_{\text{F1}}$    & 69.3 & 73.0 & 69.9 & 69.1 \\
ScienceQA             & 68.1 & 68.6 & 68.6 & 68.0 \\
MME                   & 1681 & 1707 & 1688 & 1677 \\
MMVP$_{\text{img}}$   & 67.0 & 68.1 & 67.2 & 66.7 \\
MMVP$_{\text{pair}}$  & 35.6 & 37.8 & 35.7 & 35.2 \\
\bottomrule
\end{tabular}
\label{tab:controls}
\end{table}

\section{Conclusion}
In this paper, we uncover a consistent decoder attention pathology in modern VLMs: early and late layers repeatedly collapse onto prompt-invariant visual regions, while mid layers remain the main locus of prompt-conditioned grounding.
Building on this diagnosis, we propose SAGE, a simple training-free inference-time method that reweights attention with a token-aligned ROI mask to prioritize relevant evidence.
SAGE can be instantiated with backbone-derived ROI masks from standard vision backbones such as CLIP, ViT, and DINOv3. More broadly, our results suggest that decoder attention allocation is not only a symptom of grounding failure, but also a useful intervention point for improving visually grounded generation.
We further show strong layer dependence and that intervening at appropriately chosen layers yields the most reliable improvements.
Across multiple VLM families and benchmarks, SAGE yields broad improvements in many evaluated settings, suggesting that correcting decoder-side attention allocation is an effective handle for reducing visually unfaithful outputs.

\section*{Limitations}

SAGE relies on token-aligned ROI masks derived from standard vision backbones, and its effectiveness depends on the quality of these estimates.
The three sources differ in what signal they encode: CLIP scores each patch by its similarity to the query embedding, while ViT and DINOv3 derive relevance from CLS-token attention and therefore capture image saliency.
SAGE benefits from both kinds of signal, which suggests that steering toward task-relevant regions does not require the mask itself to be query-conditioned; extending saliency-based sources with query conditioning is a natural direction for future work.
As illustrated in Figure~\ref{fig:failure_roi_instability}, fragmented or misaligned ROI estimates can steer attention toward irrelevant visual tokens and suppress the true evidence.
The random-ROI control in Section~\ref{sec:exp_controls} bounds the other end of this dependence, since an uninformative mask of the same size and bias strength removes the gain entirely.
A deployed system would therefore benefit from a confidence signal that decides when to intervene, for instance a margin on the patch relevance scores; we leave this, together with improved patch-to-token alignment, to future work.

The intervention itself is deliberately simple.
We apply one global $(\alpha,\beta)$ across all models and benchmarks to avoid per-task tuning, and the mask is held fixed across decoding steps and broadcast to all attention heads.
This keeps SAGE easy to deploy and reproduce, but it also emphasizes the same region for every generated token, which suits questions with a single decisive region better than multi-region descriptions or multi-step reasoning.
Per-step masks, head-selective steering, and per-model tuning of the bias strength are all compatible with the formulation in Eq.~\eqref{eq:sage_logit_intervention}, as is extending the single-image mask to multi-image and video.

Two aspects of the diagnosis remain open.
SAGE is sink-aware at the level of depth rather than of individual tokens: the PIS diagnostics identify where in the decoder steering is useful, and the additive bias then reallocates the attention budget through Eq.~\eqref{eq:odds_ratio}.
This design needs no per-token sink classifier, but whether explicitly suppressing sink tokens would add to the depth-level effect is untested, as is whether the same normalized depth transfers across model families without re-running the diagnostic.
Our characterization of prompt-invariant sinks is likewise empirical: the structure is consistent across models and across two independent diagnostics, but we do not explain why early and late layers develop it while mid layers remain prompt-responsive, and we regard this formation mechanism as an interesting open problem.

\section*{Ethical Considerations}

This work studies decoder-side attention steering in vision-language models and aims to improve faithfulness to visual evidence.
Although improved grounding may reduce visually unsupported outputs, the proposed method does not guarantee correctness and may still fail when ROI masks are inaccurate or unstable.
Accordingly, SAGE should not be treated as a safety mechanism in high-stakes settings, and human oversight and task-specific validation remain necessary before deployment.
Because the method operates on image inputs that may contain sensitive content, practitioners should also follow appropriate privacy and data-governance practices in downstream use.

\section*{Acknowledgment}
This research was supported by the AI Seoul Tech Research Support Program of the Seoul Future Foundation.
This work was also supported by the Institute of Information \& Communications Technology Planning \& Evaluation (IITP) grant funded by the Korea government (MSIT) [RS-2021-II211341, Artificial Intelligence Graduate School Program (Chung-Ang University)].

\bibliography{custom}
\clearpage

\appendix

\section{Additional Implementation Details}
\label{app:impl}

\subsection{Hyperparameters and Training Hardware Settings}
\label{app:settings}

Table~\ref{tab:impl_settings} summarizes the key configuration choices used in our SAGE implementation.
Unless otherwise noted, these settings are shared across the experiments reported in the main paper.
We use $(\alpha,\beta)=(5,2)$ as the default bias strengths unless otherwise stated.
For all models we use deterministic greedy decoding
($\texttt{do\_sample}=\texttt{False}$) for every arm, including baselines and
comparison methods, and otherwise follow the standard decoding settings
recommended by each model family. All experiments are run on NVIDIA GPUs (A6000) using FP16 mixed precision.

\begin{table*}[h]
\centering
\caption{\textbf{Key hyperparameters and settings for SAGE.}
We use $(\alpha,\beta)=(5,2)$ as the default bias strengths unless otherwise stated.}
\label{tab:impl_settings}
\vspace{1mm}
\small
\begin{tabular}{p{3.2cm} | p{9.2cm}}
\toprule
\textbf{Component} & \textbf{Value / Description} \\
\midrule
Local VLM &
Vision-encoder + decoder-only LLM VLMs (e.g., LLaVA-1.5~\cite{liu2024improved}/Qwen2-VL~\cite{wang2024qwen2}/InternVL3~\cite{zhu2025internvl3}/NVILA~\cite{liu2025nvila}) with standard preprocessing and decoding settings. \\
\midrule
ROI sources &
Patch-level ROI masks derived from standard pretrained vision backbones (CLIP/ViT/DINOv3). \\
\midrule
Patch relevance scoring &
Backbone-specific patch relevance scores are thresholded to form a binary ROI mask; the same scoring rule is used consistently within each source. \\
\midrule
Patch-to-token mapping &
The patch-level mask is aligned to the visual-token grid used by the evaluated VLM; the mapping is fixed per model and preprocessing configuration. \\
\midrule
Patch grid &
Model-specific patch grid size determined by the vision encoder and preprocessing; grid configuration is fixed per model and recorded in the experiment scripts. \\
\midrule
Intervention layers &
Layer set $\mathcal{L}_{\mathrm{int}}$ specified as a list or range (single-layer or multi-layer), fixed per experiment. \\
\midrule
Bias strengths &
Default $(\alpha,\beta)=(5,2)$ for ROI amplification and non-ROI suppression in Eq.~\eqref{eq:sage_logit_intervention}. \\
\midrule
Decoding &
Greedy \texttt{do\_sample = False}; maximum new tokens set per model family; decoding settings are held fixed within each benchmark. \\
\bottomrule
\end{tabular}
\end{table*}

\subsection{Intervention Scope and Implementation}
\label{app:hooking}

SAGE is implemented as a training-free, inference-time intervention that modifies attention allocation inside selected decoder layers without updating model parameters.
In our implementation, we instrument the self-attention modules of target layers $\ell \in \mathcal{L}_{\mathrm{int}}$ so that the additive bias in Eq.~\eqref{eq:sage_logit_intervention} is applied during attention computation.

\textbf{Scope.}
The intervention uses a token-aligned ROI mask $\mathbf{m}\in\{0,1\}^{N}$ and applies the additive bias only to attention logits corresponding to visual-token key positions; text-token keys remain unchanged.
Within each selected layer, the same ROI mask is broadcast to all heads and all query positions in the layer's attention operation.
All settings are kept fixed within each experiment configuration.

\textbf{Layer configuration.}
The intervention layer set $\mathcal{L}_{\mathrm{int}}$ supports both single-layer and multi-layer modes and is specified by a layer index list or range in the experiment scripts (e.g., ``0'', ``0,5,10'', ``0--5'').

\begin{algorithm}[h]
\caption{Implementation sketch of SAGE instrumentation (inference-time)}
\label{alg:sage_instrumentation}
\begin{algorithmic}[1]
\Require Model $M$, ROI mask $\mathbf{m}$, layer set $\mathcal{L}_{\mathrm{int}}$, strengths $\alpha,\beta$
\State Configure $M$ to apply Eq.~\eqref{eq:sage_logit_intervention} in all $\ell \in \mathcal{L}_{\mathrm{int}}$
\State $Y \gets M.\mathrm{generate}(I,Q)$
\State \Return $Y$
\end{algorithmic}
\end{algorithm}

\subsection{ROI Mask Construction Details}
\label{app:roi_details}

We summarize the local ROI construction pipeline used in our main experiments and the alignment from patch-level relevance maps to token-aligned ROI masks.

\textbf{Local backbone-derived ROI masks.}
We instantiate SAGE with ROI masks derived from standard vision backbones such as CLIP, ViT, and DINOv3.
For each image--query pair, a source-specific patch-level relevance score is computed and thresholded to obtain a binary patch mask.

\textbf{Patch-to-token alignment.}
The binary patch mask is then aligned to the visual-token grid used by each evaluated VLM.
Although the effective grid size can differ across model families depending on the vision encoder and preprocessing pipeline, the mapping from patch-level relevance to token-aligned mask is fixed within each experiment.

\textbf{Model-specific patch grids.}
Different VLM families may use different effective patch grids depending on the vision encoder and image preprocessing.
We align the patch-level ROI mask to the corresponding model's visual-token grid using the model's configured grid size and release the exact settings in the experiment scripts.

\subsection{Benchmark Protocols and Evaluation Prompts}
\label{app:bench_protocols}

We follow the standard evaluation protocols released by each benchmark and keep decoding settings fixed within each benchmark across all compared methods.
Below we briefly summarize what each benchmark probes in VLMs and the primary metric we report.

\textbf{TextVQA.}~\cite{singh2019towards}
OCR-centric VQA that tests whether a model can read scene text and ground it to answer questions.
We report VQA-style exact-match accuracy.

\textbf{POPE.}~\cite{li2023evaluating}
A hallucination/faithfulness diagnostic using yes/no questions about object existence.
We report accuracy and F1 following the official protocol.

\textbf{GQA.}~\cite{hudson2019gqa}
A compositional visual reasoning benchmark that probes object/attribute/relation reasoning with grounding.
We report standard accuracy on the evaluated split.

\textbf{ScienceQA.}~\cite{lu2022learn}
A multimodal multiple-choice benchmark for scientific reasoning requiring integration of perception and reasoning.
We report multiple-choice accuracy.

\textbf{MME.}~\cite{fu2026mme}
A broad diagnostic suite covering perception (e.g., OCR, count, color) and cognition (e.g., commonsense, numerical reasoning).
We report the aggregated score under the official protocol.

\textbf{MMVP.}~\cite{tong2024eyes}
A paired-image visual pattern benchmark that stresses sensitivity to subtle visual differences and discourages language priors.
We report strict pair accuracy (a pair is correct only if both questions are answered correctly).

\textbf{MMVet.}~\cite{yu2023mm}
An integrated capability benchmark with open-ended questions scored by an LLM-based evaluator.
We report the resulting score following the benchmark evaluator protocol.

\noindent\textbf{Evaluation prompts.}
We use each benchmark's recommended input formatting and evaluation scripts; any additional prompt templates used in our runs are included in the released supplementary materials.

\noindent\textbf{Artifact use.}
We use publicly released research artifacts, including model checkpoints, vision backbones, benchmark datasets, and official evaluation protocols, under their respective licenses and terms.
We use these artifacts only for research evaluation and do not redistribute the underlying datasets or model weights.

\subsection{Detailed Algorithm}
\label{app:algorithm_sage}

Algorithm~\ref{alg:sage} summarizes the full SAGE pipeline. Given an input image $I$, a query $Q$, and a local VLM $M$ (vision encoder + decoder-only LLM), SAGE performs a training-free intervention during generation by selectively biasing the logits of visual tokens at a chosen set of decoder layers $\mathcal{L}_{\mathrm{int}}$.

\paragraph{Step 1: ROI mask construction.}
SAGE first constructs a binary ROI mask $\mathbf{m}$ over the visual tokens produced by the vision encoder:
\[
\mathbf{m} \leftarrow \textsc{GetROIMask}(I, Q, \mathcal{S}).
\]
Here, the ROI source $\mathcal{S}$ is a local vision backbone such as CLIP, ViT, or DINOv3.
For the selected source, we compute a patch-level relevance map, threshold it to obtain a binary patch mask, and align the result to the visual-token grid of the evaluated VLM.
The output mask $\mathbf{m}\in\{0,1\}^{N_v}$ marks which visual tokens belong to the guided ROI.

\paragraph{Step 2: Enable logit-level steering at selected layers.}
Next, SAGE activates a lightweight forward hook that applies Eq.~\eqref{eq:sage_logit_intervention} to the visual-token logits at each layer $\ell\in\mathcal{L}_{\mathrm{int}}$:
\[
\textsc{EnableSteering}(M, \mathcal{L}_{\mathrm{int}}, \mathbf{m}, \alpha, \beta).
\]
Intuitively, $\alpha$ controls the strength of encouraging ROI tokens, while $\beta$ controls the strength of suppressing non-ROI tokens (as defined in Eq.~\eqref{eq:sage_logit_intervention}). Importantly, this intervention does not require gradient updates or any modification to model parameters; it only alters the logits used to form attention over visual tokens during decoding. By biasing visual-token logits, this intervention implicitly reshapes the attention distribution over visual tokens during decoding, without any parameter updates.

\paragraph{Step 3: Generate with SAGE enabled.}
With steering enabled, the model generates the answer as usual:
\[
Y \leftarrow M.\texttt{generate}(I, Q).
\]
All other components of decoding (e.g., sampling strategy, max tokens, temperature) remain unchanged unless specified elsewhere.

\paragraph{Step 4: Disable steering and return.}
Finally, SAGE removes the hook to restore the original model behavior:
\[
\textsc{DisableSteering}(M),
\]
and returns the generated output $Y$. This explicit enable/disable design ensures SAGE is modular and can be applied on a per-query basis without persistent side effects.

\paragraph{Practical notes.}
In practice, $\mathcal{L}_{\mathrm{int}}$ can be chosen as a single layer or a small set of layers to trade off effectiveness and overhead. The only additional per-query cost comes from computing $\mathbf{m}$ and applying a constant-time logit bias at the selected layers; no extra training or finetuning is required.

\begin{algorithm}[t]
\caption{SAGE: Local ROI-Guided, Training-Free Attention Steering}
\label{alg:sage}
\begin{algorithmic}[1]
\Require Image $I$, Query $Q$, local VLM $M$ (vision encoder + decoder-only LLM)
\Require ROI source $\mathcal{S}$ (e.g., CLIP/ViT/DINOv3 patch mask)
\Require Intervention layer set $\mathcal{L}_{\mathrm{int}}$, hyperparameters $\alpha,\beta$
\Ensure Answer $Y$
\State $\mathbf{m} \gets \Call{GetROIMask}{I, Q, \mathcal{S}}$
\State \Call{EnableSteering}{$M, \mathcal{L}_{\mathrm{int}}, \mathbf{m}, \alpha, \beta$}
\Comment{apply Eq.~\eqref{eq:sage_logit_intervention} to visual-token logits at selected layers}
\State $Y \gets M.\mathrm{generate}(I, Q)$
\State \Call{DisableSteering}{$M$}
\State \Return $Y$
\end{algorithmic}
\end{algorithm}

\subsection{Hyperparameters and Sensitivity}
SAGE is controlled by two scalar hyperparameters in Eq.~(\ref{eq:sage_logit_intervention}):
$\alpha$, the ROI-side amplification, and $\beta$, the non-ROI-side
suppression. Table~\ref{tab:sensitivity_ablation} reports a seven-point sensitivity study over
$(\alpha,\beta)$ on MMVP with the ROI source and intervention layer held
fixed. Performance is not flat across settings: one-sided variants
(pull-only, push-only) and overly aggressive bias both fall well below the
default, spanning roughly nine points. We therefore fix
$(\alpha,\beta)=(5,2)$ globally across all models and benchmarks rather than
tuning per task, and treat the bias strength as a parameter that must be kept
moderate rather than one that can be pushed for larger gains.

\begin{table}[h]
\centering
\small
\caption{\textbf{Sensitivity of SAGE to logit-level intervention strength $(\alpha,\beta)$.}
We report MMVP (\%).}
\begin{tabular}{lccc}
\toprule
Variant & $(\alpha,\beta)$ & MMVP (\%) \\
\midrule
No re-weighting & $(0,0)$ & 61.48\\
Pull-only       & $(0,2)$ & 54.07\\
Push-only       & $(5,0)$ & 57.49\\
Weak            & $(3,1)$ & 61.91\\
\textbf{Default}& $(5,2)$ & \textbf{62.96}\\
Strong          & $(7,3)$ & 59.39\\
Aggressive      & $(9,4)$ & 55.93\\
\bottomrule
\end{tabular}
\label{tab:sensitivity_ablation}
\end{table}

When reporting sensitivity, we recommend fixing all other settings (ROI source, layer set, decoding parameters) and varying only one parameter at a time.

\section{Additional Layer-wise Analyses}
\label{app:layerwise}

\subsection{Per-benchmark Single-layer Sweep Curves}
\label{app:sweep_curves}

Figures~\ref{fig:roi_source_clip}--\ref{fig:roi_source_dinov3} provide per-benchmark single-layer sweep curves for LLaVA-1.5 using local ROI masks derived from CLIP, ViT, and DINOv3, respectively.
For each decoder layer index $\ell$, we apply SAGE to a single layer $\mathcal{L}_{\mathrm{int}}=\{\ell\}$ and report the resulting benchmark score using the official evaluation scripts.
Across sources and tasks, the response to intervention is clearly layer-dependent: some layers yield consistent gains over the no-intervention baseline, while others are neutral or occasionally degrade performance.
These curves complement the aggregated summaries in the main paper and show that the effectiveness of steering is not tied to a single local ROI source.

\subsection{Exact Best-Layer Indices for Main Results}
\label{app:best_layers}

For transparency, we list in Table~\ref{tab:best_layers_main} the exact single-layer indices used to produce the \textbf{Best} results in Table~\ref{tab:main_results}.
These indices are provided as supplementary detail; in the main text, cross-model discussion uses normalized decoder depth rather than raw layer number.

\begin{table*}[t]
\centering
\small
\caption{\textbf{Exact model-specific best-layer indices used for the \textbf{Best} results in Table~\ref{tab:main_results}.}
Indices are raw layer numbers within each model and should not be compared directly across architectures; cross-model discussion in the main text uses normalized decoder depth.}
\label{tab:best_layers_main}
\setlength{\tabcolsep}{4.0pt}
\resizebox{\textwidth}{!}{
\begin{tabular}{l c c c c c c c c c c}
\toprule
\textbf{Model} & \textbf{TextVQA} & \textbf{POPE} & \textbf{GQA} & \textbf{SQA} & \textbf{MME} & \textbf{MMVet} & \textbf{MMVP Img} & \textbf{MMVP Pair} & \textbf{MMVP-O Img} & \textbf{MMVP-O Pair} \\
\midrule

\multicolumn{11}{l}{\textit{LLaVA 1.5-13B}} \\
+ SAGE (CLIP)   & L17 & L3 & L20 & L18 & L24 & L20 & L26 & L26 & L18 & L6  \\
+ SAGE (ViT)    & L29 & L10 & L19 & L20 & L22 & L18 & L18 & L18 & L2 & L18 \\
+ SAGE (DINOv3) & L30 & L10 & L30 & L25 & L16 & L20 & L26 & L26 & L18 & L18  \\
\midrule

\multicolumn{11}{l}{\textit{Qwen2-VL-7B}} \\
+ SAGE (CLIP)   & L11 & L18 & L21 & L15 & L25 & L14 & L18 & L18 & L15 & L15 \\
+ SAGE (ViT)    & L13 & L18 & L15 & L15 & L20 & L17 & L27 & L27 & L15 & L15 \\
+ SAGE (DINOv3) & L16 & L15 & L18 & L18 & L27 & L14 & L18 & L18 & L12& L12 \\
\midrule

\multicolumn{11}{l}{\textit{InternVL3-2B}} \\
+ SAGE (CLIP)   & L11 & L18 & L21 & L15 & L25 & L25 & L18 & L18 & L8 & L8 \\
+ SAGE (ViT)    & L15 & L18 & L14 & L19 & L24 & L17 & L27 & L27 & L16 & L16 \\
+ SAGE (DINOv3) & L9 & L18 & L27 & L19 & L27 & L15 & L21 & L21 & L13 & L13 \\
\midrule

\multicolumn{11}{l}{\textit{NVILA-8B}} \\
+ SAGE (CLIP)   & L19 & L20 & L12 & L15 & L15 & L10 &  L10 & L10 & L20 & L20  \\
+ SAGE (ViT)    & L13 & L20 & L18 & L17 & L16 & L12 & L15 & L15 & L13 & L13  \\
+ SAGE (DINOv3) & L10 & L10 & L15 & L18 & L5 & L21 & L15 & L15 & L20 & L20 \\
\bottomrule
\end{tabular}
}
\end{table*}

\subsection{Additional Multi-layer Set Comparisons}
\label{app:multilayer_more}

To complement the group-level summaries in Table~\ref{tab:depth_group_summary}, we provide a small number of illustrative exact layer combinations for the multi-layer setting.
Rather than exhaustively listing every tested set, we report selected examples that reflect the stronger non-early compositions and a few failure-prone early-layer cases.

Tables~\ref{tab:2layer_examples}--\ref{tab:failure_examples} summarize illustrative 2-layer and 3-layer configurations, together with selected degradation cases.
Across these examples, sets concentrated on mid and/or final depths tend to produce stronger performance than sets involving the earliest layers.
These examples are consistent with the depth sensitivity observed in the main paper and help ground the group-level averages with concrete layer indices.

\begin{table}[h]
\centering
\small
\caption{\textbf{Illustrative 2-layer SAGE sets on TextVQA (LLaVA-1.5).}
These examples are selected to represent stronger non-early depth compositions.}
\label{tab:2layer_examples}
\begin{tabular}{l l c}
\toprule
\textbf{Layer set} & \textbf{Depth composition} & \textbf{Accuracy (\%)} \\
\midrule
(14,16) & Mid--mid     & 62.43 \\
(17,21) & Mid--final   & 62.72 \\
(28,31) & Final--final & 62.58 \\
\bottomrule
\end{tabular}
\end{table}

\begin{table}[h]
\centering
\small
\caption{\textbf{Illustrative 3-layer SAGE sets on TextVQA (LLaVA-1.5).}
These examples are selected to represent stronger multi-layer compositions away from the earliest decoder layers.}
\label{tab:3layer_examples}
\begin{tabular}{l l c}
\toprule
\textbf{Layer set} & \textbf{Depth composition} & \textbf{Accuracy (\%)} \\
\midrule
(14,18,21) & Mid $\times 3$         & 62.33 \\
(17,24,30) & Mid--final--final      & 62.63 \\
(24,27,30) & Final $\times 3$       & 62.53 \\
\bottomrule
\end{tabular}
\end{table}

\begin{table}[!htbp]
\centering
\small
\caption{\textbf{Illustrative failure-prone early-layer sets on TextVQA (LLaVA-1.5).}
These examples show that sets concentrated near the earliest decoder layers can still lead to substantial degradation.}
\label{tab:failure_examples}
\begin{tabular}{l l c}
\toprule
\textbf{Layer set} & \textbf{Depth composition} & \textbf{Accuracy (\%)} \\
\midrule
(0,2)   & Very-early only       & 12.07 \\
(0,2,4) & Very-early $\times 3$ & 7.12 \\
(1,2,3) & Very-early $\times 3$ & 1.12 \\
\bottomrule
\end{tabular}
\end{table}

\subsection{Layer Skip Ablation}
\label{app:skip_ablation}

To further probe depth-wise roles of the decoder, we perform a layer-skip ablation on LLaVA-1.5 that removes computation from selected decoder layers at inference time.
For a chosen skip set $\mathcal{L}_{\mathrm{skip}}$, we bypass each layer $\ell \in \mathcal{L}_{\mathrm{skip}}$ by forwarding its input hidden states to the next layer without applying the layer's transformation.
This ablation adds no new parameters and isolates how sensitive the model's final behavior is to perturbations at different decoder depths.

\begin{figure}[t]
\centering
\includegraphics[width=\linewidth]{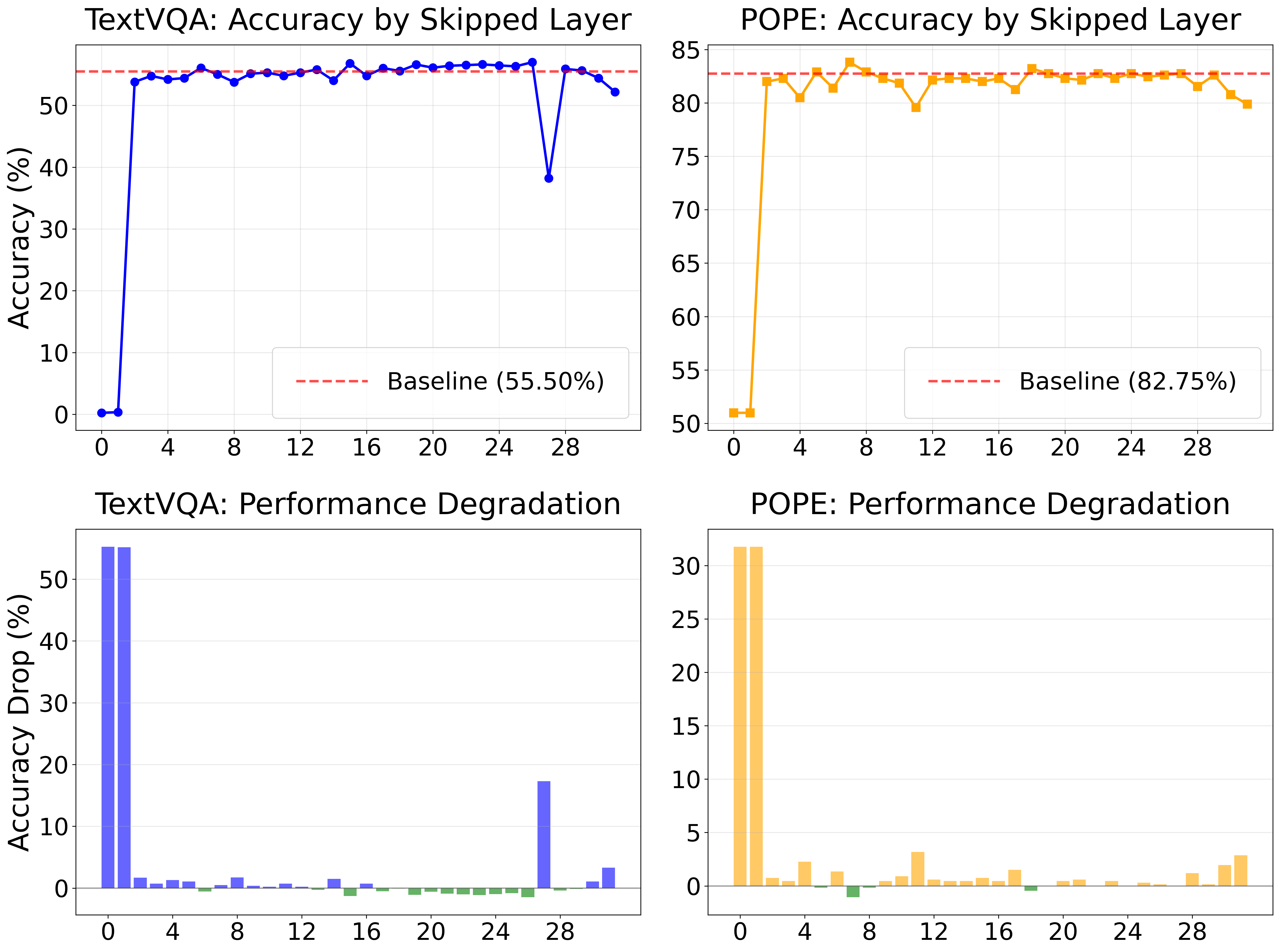}
\caption{\textbf{Layer-skip ablation on LLaVA-1.5.}
For each decoder layer, we skip only that layer at inference and evaluate performance.
The top row reports TextVQA and POPE accuracy versus the skipped depth, and the bottom row shows the change relative to the no-skip baseline.}
\label{fig:skip_ablation}
\end{figure}

Figure~\ref{fig:skip_ablation} reports the impact of skipping different layers on downstream performance.
Skipping is not uniformly tolerated across depth: certain layers lead to more noticeable degradation than others.
This provides complementary evidence to the single-layer intervention sweep in the main paper and supports the interpretation that some early decoder layers can be more sensitive to perturbations.

\section{ROI Sources from Standard Vision Backbones}
\label{app:roi_sources}

SAGE only requires a token-aligned binary mask $\mathbf{m}\in\{0,1\}^N$ over visual tokens.
In this section, we summarize the three ROI sources used in our study: CLIP, ViT, and DINOv3.
These sources produce different patch-level relevance patterns, but the intervention in Eq.~\eqref{eq:sage_logit_intervention} is identical across sources; only the procedure that produces $\mathbf{m}$ changes.

\section{Additional Controlled Comparisons}
\label{app:controls_extra}

This appendix collects two comparisons that share the fixed-protocol setting of
Section~\ref{sec:exp_controls} but are not needed to follow the main argument.
Both use no benchmark-specific selection of the intervention layer.

\subsection{Comparison with an Inference-Time Decoding Baseline}
\label{app:vcd}

Section~\ref{sec:exp_controls} compares the fixed-layer setting against a
re-measured baseline and a matched random-ROI control, but not against another
published intervention.
To place SAGE relative to existing inference-time methods, we run visual
contrastive decoding (VCD~\cite{leng2024mitigating}) under exactly the same protocol with its official hyperparameters
($\alpha=1.0$, $\beta=0.1$, noise step $500$).
VCD reaches $50.8$ on TextVQA and $59.5$ on GQA, and is neutral-to-negative on
POPE ($-0.2$) and ScienceQA ($-0.3$).
Fixed-layer SAGE reaches $54.6$ and $62.3$ on the same two benchmarks
(Table~\ref{tab:controls}).
The two methods therefore differ not only in magnitude on the
localized-evidence benchmarks but also in whether they are neutral elsewhere.
This comparison covers a single method on a single model family, and we do not
read it as a general ranking of inference-time interventions.

\subsection{Fixed-Protocol Control Across Model Families}
\label{app:fixed_protocol}

The controls in Section~\ref{sec:exp_controls} are established on
LLaVA-1.5.
To check that the fixed-protocol setting is not specific to that model, we
repeat it across four families (LLaVA-1.5, Qwen2-VL, InternVL3, NVILA) using
one fixed target layer per family, fixed $(\alpha,\beta)=(5,2)$, a fixed top-48
mask, and no benchmark-specific selection at any point.
Under this setting the same push--pull intervention yields moderate gains:
GQA $+2.55$, TextVQA $+1.58$, and MMVP-pair $+6.67$.
We report them as evidence that the intervention is not confined to one
architecture, not as a like-for-like comparison with the main table.

\subsection{ROI Sources}
\label{app:roi_sources_list}

\textbf{Backbone-derived ROI (CLIP/ViT/DINOv3).}
We derive ROI masks from standard pretrained vision backbones without explicit detector-based grounding.
For each source, we compute patch-level relevance scores and threshold them to obtain a binary patch mask, which is then aligned to the visual-token grid used by the evaluated VLM.
Although the resulting masks can differ in sharpness and spatial coverage across sources, all three provide usable local guidance signals for SAGE.

\subsection{Single-layer Sweep with Backbone-Derived ROI Masks}
\label{app:roi_source_ablation}

\begin{figure*}[p]
    \centering
    \includegraphics[width=\textwidth]{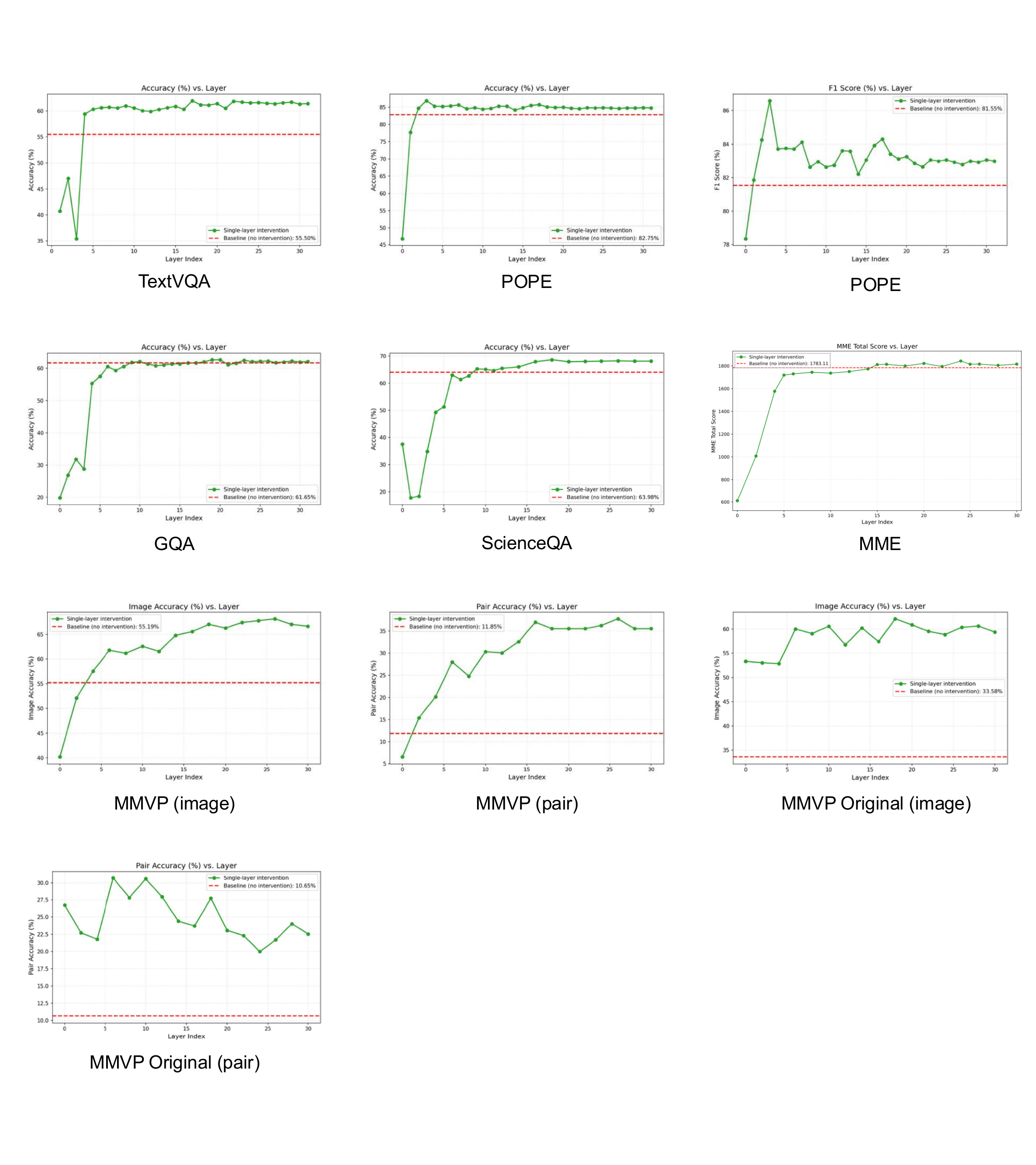}
    \caption{\textbf{Single-layer sweep with a CLIP-derived ROI mask.}
    We report per-benchmark performance as a function of the intervened layer using a CLIP-based patch-level ROI mask.}
    \label{fig:roi_source_clip}
\end{figure*}

\begin{figure*}[p]
    \centering
    \includegraphics[width=\textwidth]{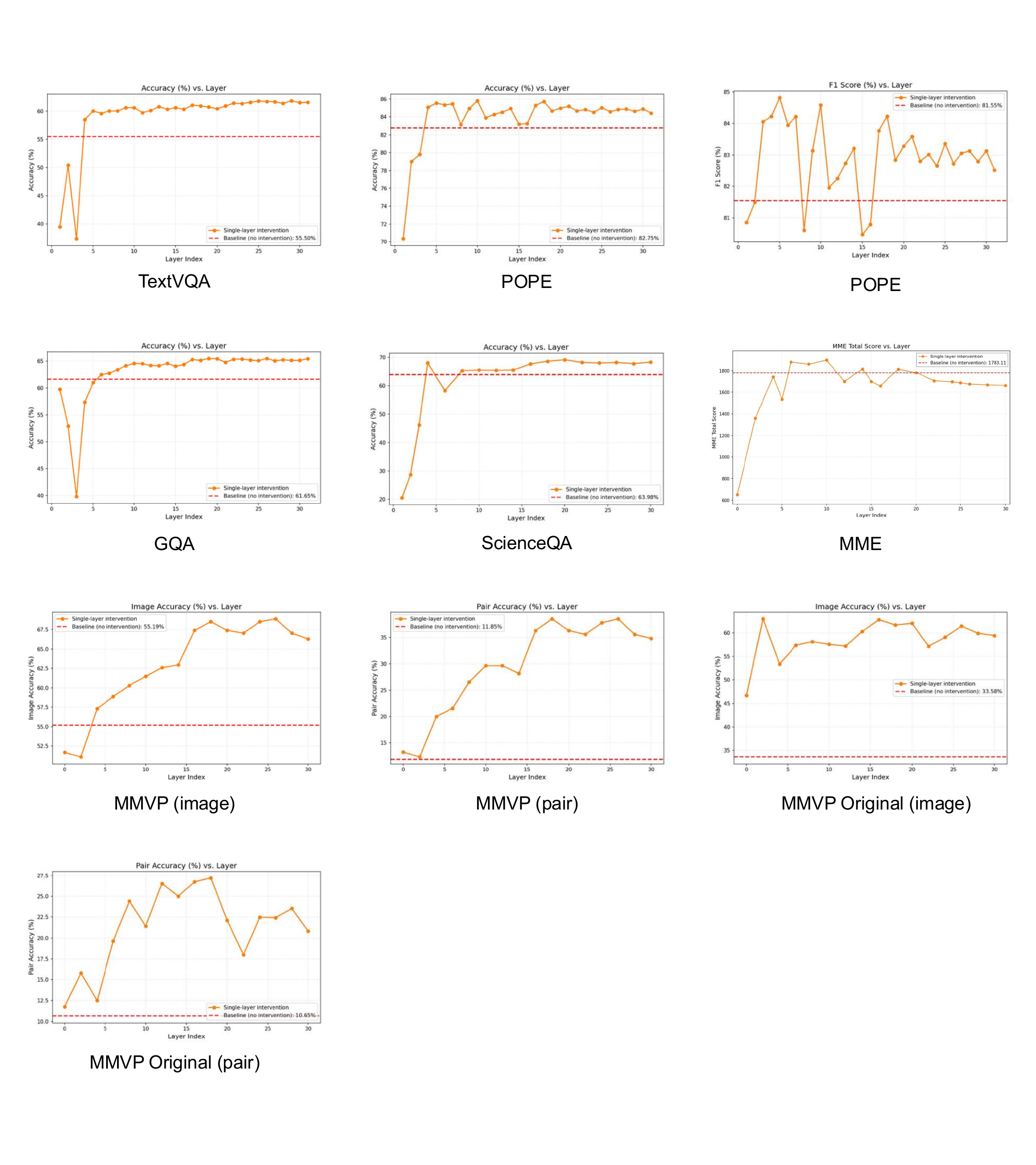}
    \caption{\textbf{Single-layer sweep with a ViT-derived ROI mask.}
    We report per-benchmark performance as a function of the intervened layer using a ViT-based patch-level ROI mask.}
    \label{fig:roi_source_vit}
\end{figure*}

\begin{figure*}[p]
    \centering
    \includegraphics[width=\textwidth]{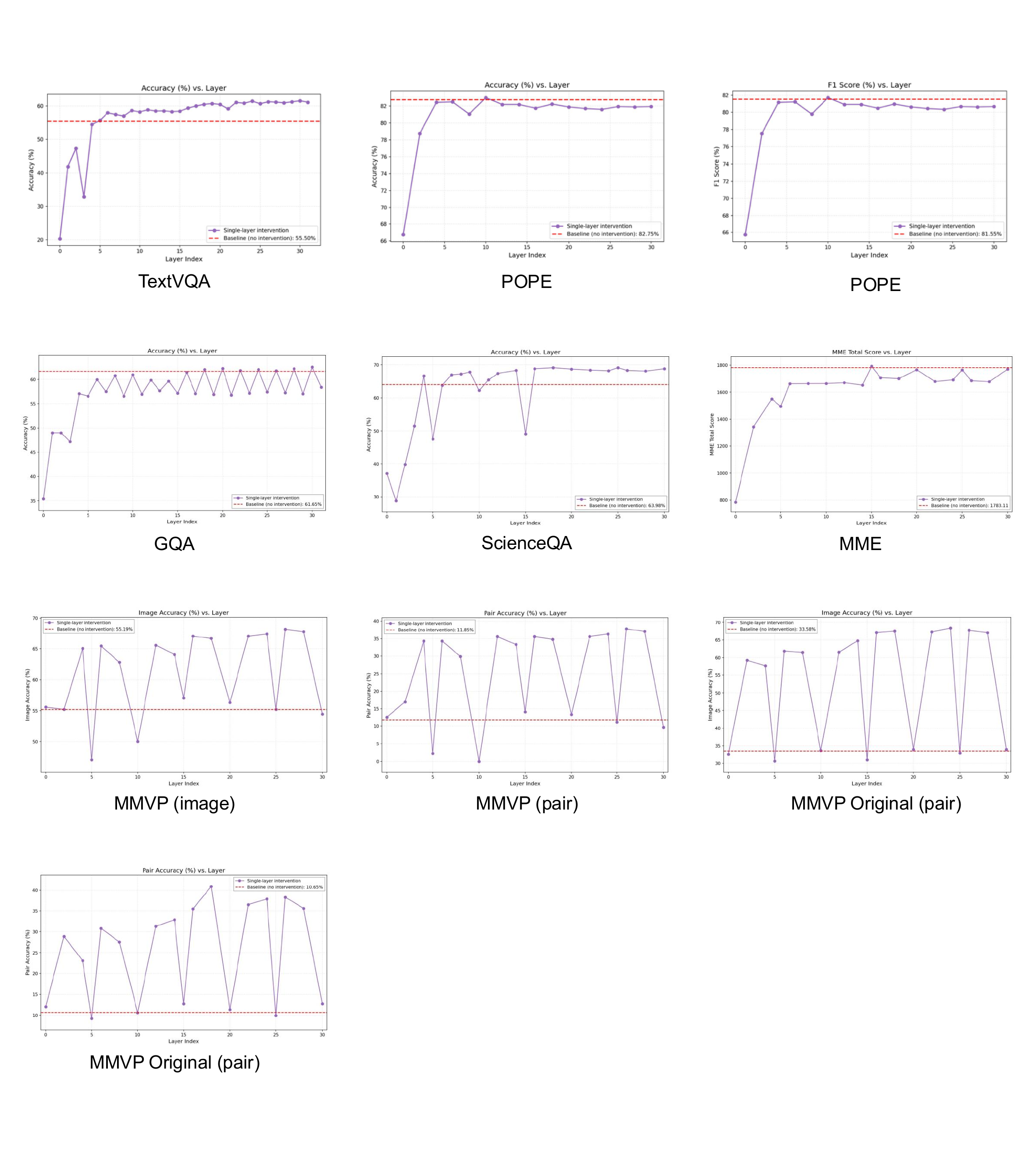}
    \caption{\textbf{Single-layer sweep with a DINOv3-derived ROI mask.}
    We report per-benchmark performance as a function of the intervened layer using a DINOv3-based patch-level ROI mask.}
    \label{fig:roi_source_dinov3}
\end{figure*}

Figures~\ref{fig:roi_source_clip}--\ref{fig:roi_source_dinov3} report additional single-layer sweep results using ROI masks derived from CLIP, ViT, and DINOv3.
These results further support that SAGE remains effective under multiple local ROI construction mechanisms.

\subsection{Qualitative Comparison of ROI Masks Across Sources}
\label{app:roi_sources_qual}

Figure~\ref{fig:failure_roi_instability} visualizes ROIs produced by different local ROI sources for the same image--query pairs.
CLIP, ViT, and DINOv3 can produce different mask shapes and degrees of spatial concentration, reflecting differences in their patch-level relevance patterns.
Despite these differences, all three sources can be converted into token-aligned masks and used by SAGE without changing the intervention rule.

\section{Mechanistic Analysis Details}
\label{app:mech}

This appendix specifies the exact definitions and computation procedures used in our mechanistic analyses (sink structure, spectral dominance, and attention-mass redistribution).

\subsection{Constructing the Logit Matrix for Spectral Analysis}
\label{app:mech_svd}

Fix a decoder layer $\ell$ and head $h$.
For each evaluation instance, we collect pre-softmax attention logits from text-query positions to visual-token keys.
Let $\mathbf{a}^{(\ell,h)}_{p,t}\in\mathbb{R}^{N}$ denote the restricted logit vector over the $N$ visual-token keys at decoding step $t$ for prompt (instance) $p$.
We stack these vectors across instances and decoding steps into a matrix
\begin{equation}
\mathbf{A}^{(\ell,h)} \in \mathbb{R}^{M \times N}, 
\qquad
\mathbf{A}^{(\ell,h)}[r,:] = \mathbf{a}^{(\ell,h)}_{p,t},
\end{equation}
where $r$ indexes the pair $(p,t)$ and $M$ is the total number of stacked vectors.
We then compute $\mathbf{A}^{(\ell,h)} = \mathbf{U}\mathbf{\Sigma}\mathbf{V}^\top$ and report spectral concentration using
\begin{equation}
\rho_k(\mathbf{A}) = \frac{\sum_{j=1}^{k}\sigma_j^2}{\sum_{j=1}^{\min(M,N)}\sigma_j^2},
\end{equation}
where $\{\sigma_j\}$ are singular values.

\subsection{Prompt-Invariant Sink Definition and Sink IoU}
\label{app:mech_sink_iou}

For each instance $p$ and decoding step $t$, we convert the visual-token attention weights into a set of sink tokens by selecting the top-$r\%$ tokens by attention mass.
We use $r=5$ throughout unless otherwise specified.
Given two sink sets $S_{p}$ and $S_{p'}$ computed from different prompts (or different queries) on the same image, we measure prompt-invariance using the intersection-over-union (IoU):
\begin{equation}
\mathrm{IoU}(S_{p}, S_{p'}) = \frac{|S_{p}\cap S_{p'}|}{|S_{p}\cup S_{p'}|}.
\end{equation}
We report mean IoU with standard deviation across prompt pairs.

\subsection{Attention-Mass Metrics: Sink Mass and ROI Mass}
\label{app:mech_mass}

Let $\mathbf{p}\in\mathbb{R}^{N}$ be attention weights over visual keys (after softmax) and let $\mathbf{m}\in\{0,1\}^{N}$ be the ROI mask.
We define ROI mass as the total attention assigned to ROI tokens:
\begin{equation}
\mathrm{Mass}_{\mathrm{ROI}}(\mathbf{p}) = \sum_{i=1}^{N} \mathbf{p}[i]\cdot \mathbf{m}[i].
\end{equation}
We define sink mass as the total attention assigned to the sink-token set $S$ (top-$r\%$ tokens by attention mass):
\begin{equation}
\mathrm{Mass}_{\mathrm{sink}}(\mathbf{p}) = \sum_{i\in S} \mathbf{p}[i].
\end{equation}
Unless stated otherwise, we use the same $r=5$ as in Appendix~\ref{app:mech_sink_iou}.

\subsection{Implementation Notes for Mechanistic Measurements}
\label{app:mech_impl}

All mechanistic measurements are computed under matched decoding settings between baseline and SAGE runs.
When reporting layer-wise results, we either fix a representative head or average the reported quantities across heads within the layer; we explicitly state the choice in each figure caption.

\subsection{Evidence Retention to the Final Layer}
\label{app:retention}

To test whether the steering effect survives to late decoding rather than
being a single-layer artifact, we measure a retention ratio: final-layer ROI
attention mass relative to its mid-layer peak. On LLaVA-1.5, averaged over
four benchmarks, guided steering raises this ratio from $0.77$ to $0.96$,
indicating reduced mid-to-final evidence loss and a change in routing rather
than a local change in one layer's heatmap.

\begin{figure*}[t]
\centering
\includegraphics[width=1\textwidth]{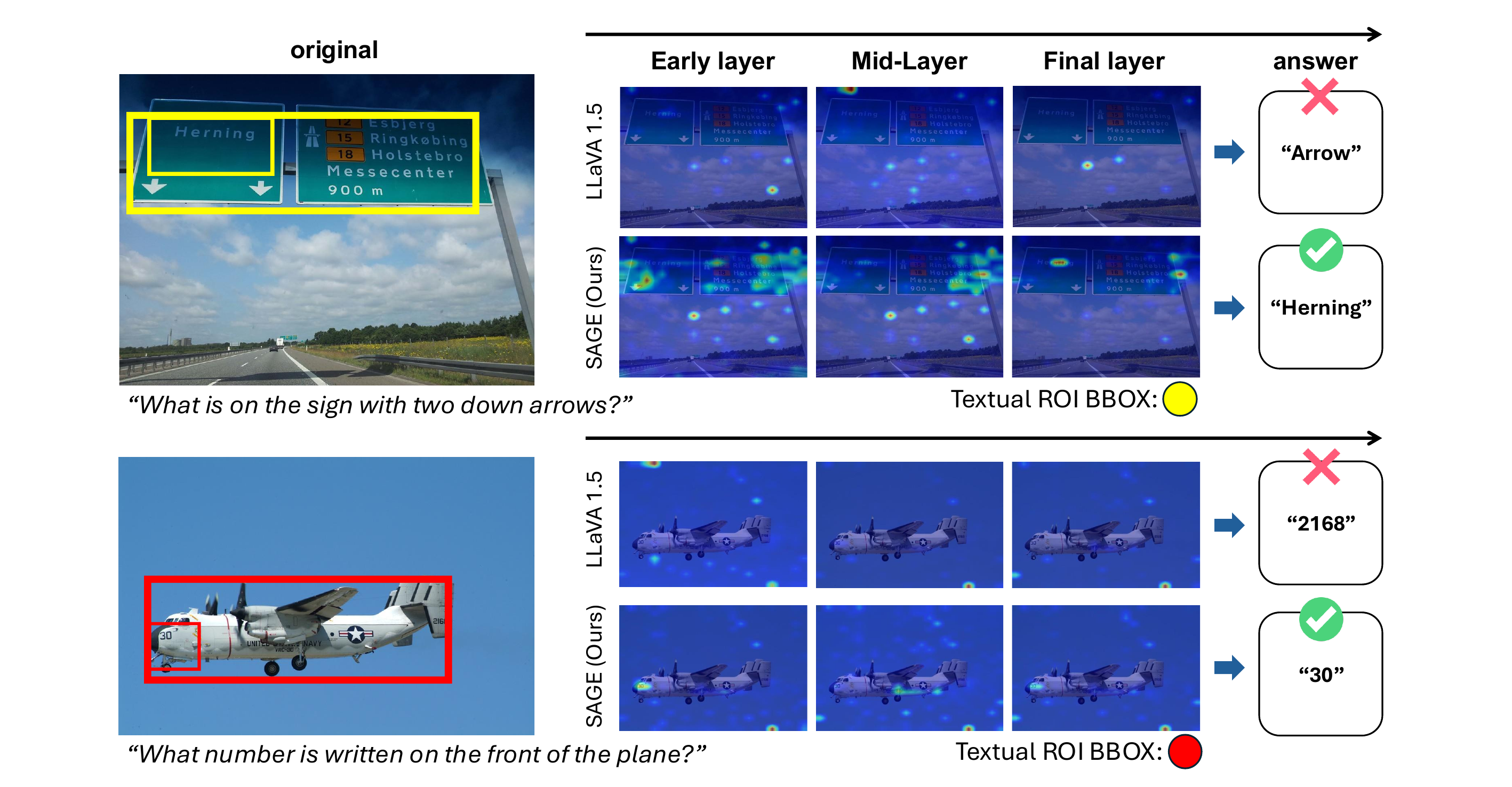} 
\caption{\textbf{Qualitative comparison of baseline vs. SAGE.}
For each example, we show the input image and query, baseline and SAGE predictions, and attention visualizations.
SAGE encourages attention to concentrate on query-relevant visual evidence, improving grounding in fine-grained cases.}
\label{fig:qualitative}
\vspace{-4mm}
\end{figure*}

\begin{figure*}[!hbt]
  \centering
  \includegraphics[width=\linewidth]{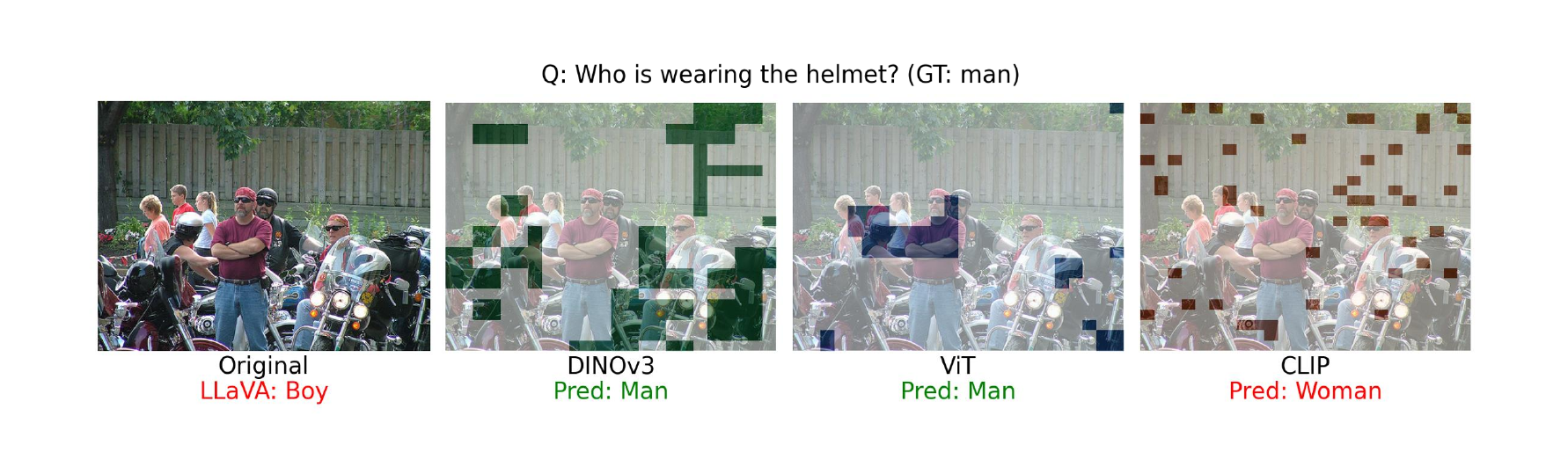}
  \caption{\textbf{Failure case under unstable local ROI estimates.}
  Different local ROI sources can produce fragmented or partially misaligned masks, which may steer attention away from the true evidence and lead SAGE to fail.}
  \vspace{-0.5em}
  \label{fig:failure_roi_instability}
\end{figure*}

\section{Qualitative Examples}
\label{app:qual}

We provide qualitative examples to illustrate how SAGE changes the model’s visual evidence usage during generation.
Figure~\ref{fig:qualitative} compares baseline predictions with SAGE under the same input image and query, together with attention visualizations over the image.
Across examples, baseline models can allocate attention to visually salient but query-irrelevant regions, which can lead to weak grounding in fine-grained questions.
In contrast, with the same ROI guidance, SAGE shifts attention toward query-relevant regions and encourages the model to rely on localized visual evidence when forming its answer.

In Figure~\ref{fig:qualitative}, SAGE is particularly helpful on queries that require fine-grained visual evidence.
For instance, when asked to read the text on a distant sign, the baseline answer can be driven by a nearby salient cue and miss the actual characters, while SAGE increases attention on the sign region and enables the correct reading.
Similarly, for the aircraft example, focusing on the front fuselage is necessary to identify the printed number; SAGE concentrates attention on this localized area and corrects the prediction.
These cases suggest that improving ROI-focused attention can be beneficial for tasks such as reading small text or verifying localized details.

\subsection{Failure Case: Sensitivity to Unstable Local ROI Estimates}
\label{app:failure_roi_instability}

Figure~\ref{fig:failure_roi_instability} shows a representative failure case where SAGE does not correct the answer when the local ROI estimate is unstable.
Since SAGE intervenes only on the region specified by the ROI source, inaccurate or fragmented local masks may steer attention toward irrelevant visual tokens and suppress the true evidence.
As a result, the model can still produce an incorrect prediction even though logit-level steering is applied.




\end{document}